\documentclass{article}
\PassOptionsToPackage{numbers,sort&compress}{natbib}

\usepackage[utf8]{inputenc} % allow utf-8 input
\usepackage[T1]{fontenc}    % use 8-bit T1 fonts
\usepackage{hyperref}       % hyperlinks
\usepackage{url}            % simple URL typesetting
\usepackage{booktabs}       % professional-quality tables
\usepackage{amsmath, amssymb, amsthm, amsfonts}       % blackboard math symbols
\usepackage{nicefrac, xfrac}       % compact symbols for 1/2, etc.
\usepackage{microtype}      % microtypography
\usepackage{xcolor}         % colors
\usepackage{bm}
\usepackage{enumitem}
\usepackage{graphicx}
\usepackage{algorithm}
\usepackage{algorithmic}
\usepackage{natbib}[numbers,sort&compress]
\usepackage{geometry}
\usepackage{authblk,textcomp}

\usepackage[cal=euler]{mathalfa}
\usepackage{libertine}
\usepackage{mathtools}

\DeclarePairedDelimiter\floor{\lfloor}{\rfloor}
\makeatletter
\newcommand{\otherlabel}[2]{\protected@edef\@currentlabel{#2}\label{#1}}
\makeatother
\newtheorem{result}{Main theoretical result}

\DeclareMathOperator*{\argmin}{arg\,min}

\DeclareMathOperator{\EX}{\mathbb{E}}% expected value
\DeclareMathOperator{\Tr}{Tr}
\newcommand{\norm}[1]{\left\lVert#1\right\rVert}

\def\mat#1{\text{#1}}
\renewcommand{\vec}[1]{\boldsymbol{#1}}

\def\mse{\text{MSE}}
\def\mmse{\text{MMSE}}
\def\sym{\text{Sym}}
\def\bo{\text{bo}}
\def\dd{\text{d}}
\def\rs{\text{rs}}
\def\amp{\text{amp}}
\def\alg{\text{alg}}
\def\it{\text{it}}

\hypersetup{pdfauthor={IdePHICS},pdftitle={Subspace},%
            colorlinks, linktocpage=true, pdfstartpage=1, pdfstartview=FitV,%
    breaklinks=true, pdfpagemode=UseNone, pageanchor=true, pdfpagemode=UseOutlines,%
    plainpages=false, bookmarksnumbered, bookmarksopen=true, bookmarksopenlevel=1,%
    hypertexnames=true, pdfhighlight=/O,%
    urlcolor=orange, linkcolor=blue, citecolor=blue
        }

\title{Subspace clustering in high-dimensions: \\ Phase transitions \& Statistical-to-Computational gap}

\author[1]{Luca Pesce}
\author[1,2]{Bruno Loureiro}
\author[1]{Florent Krzakala}
\author[3]{Lenka Zdeborov\'a}
\affil[1]{\small Ecole Polytechnique F\'{e}d\'{e}rale de Lausanne (EPFL). 
Information, Learning and Physics (IdePHICS) lab. \newline CH-1015 Lausanne, Switzerland.}
\affil[2]{\small Département d’Informatique, École Normale Supérieure - PSL \& CNRS, 45 rue d’Ulm, \newline F-75230 Paris cedex 05, France.}
\affil[3]{\small Ecole Polytechnique F\'{e}d\'{e}rale de Lausanne (EPFL).
Statistical Physics of Computation (SPOC) lab. \newline CH-1015 Lausanne, Switzerland.}
\affil[ ]{\textit {\{luca.pesce, bruno.loureiro, florent.krzakala, lenka.zdeborova\}@epfl.ch}}

\date{}

\begin{document}

\maketitle

%%%%%%%%%%%%%%%%%%%%%%%%%%%%%%%%%%%%%%%%%%%%
\begin{abstract}
A simple model to study subspace clustering  is the high-dimensional $k$-Gaussian mixture model where the cluster means are sparse vectors. Here
we provide an exact asymptotic characterization of the statistically optimal reconstruction error in this model in the high-dimensional regime with extensive sparsity, i.e. when the fraction of non-zero components of the cluster means $\rho$, as well as the ratio $\alpha$ between the number of samples and the dimension are fixed, while the dimension diverges. We identify the information-theoretic threshold below which obtaining a positive correlation with the true cluster means is statistically impossible. Additionally, we investigate the performance of the approximate message passing (AMP) algorithm analyzed via its state evolution, which is conjectured to be optimal among polynomial algorithm for this task.
We identify in particular the existence of a statistical-to-computational gap between the algorithm that require a signal-to-noise ratio $\lambda_{\text{alg}} \ge k  / \sqrt{\alpha} $ to perform better than random, and the information theoretic threshold at $\lambda_{\text{it}} \approx \sqrt{-k \rho \log{\rho}}  / \sqrt{\alpha}$.  
Finally, we discuss the case of sub-extensive sparsity $\rho$ by comparing the performance of the AMP with other sparsity-enhancing algorithms, such as sparse-PCA and diagonal thresholding.
\end{abstract}

%%%%%%%%%%%%%%%%%%%%%%%%%%%%%%%%%%%%%%%%%%%%
\section{Introduction}
\label{sec:main:intro}

With the growing size of modern data, clustering techniques play an important role in reducing the dimensionality of the features used in modern Machine Learning pipelines. Indeed, in many tasks of interest ranging from DNA sequence analysis to image classification, the relevant features are known to live in a lower-dimensional space (intrinsic dimension) than their raw acquisition format (extrinsic dimension) \cite{IDIM}. In these cases, identifying these features can help saving computational resources while significantly improving on learning performance. But given a corrupted embedding of low-dimensional features in a high-dimensional space, is it always \emph{statistically possible} to retrieve them? And if yes - how can reconstruction be achieved \emph{efficiently} in practice? In this manuscript we address these two fundamental questions in a simple model for subspace clustering: a $k$-cluster Gaussian mixture model with sparse centroids. In this model, the low-dimensional hidden features are given by the sparse centroids, which are embedded in a higher dimensional space and corrupted by additive Gaussian noise. We assume that the number of non-zero components of the centroids as well as the number of samples scales linearly with the dimension of the embedding space. Given a finite sample from the mixture, the goal of the statistician is to cluster the data, i.e. estimate the centroids (or features) as well as possible.

%%%%%%%%%%%%%%%%%%%%%%%%%%%%%%%%%%%%%%%%%%%%
\section{Model \& setting}
\label{sec:main:model}
Let $\vec{x}_{\nu}\in\mathbb{R}^{d}$, $\nu\in[n] \coloneqq \{1,\cdots, n\}$ denote $n$ i.i.d. data points drawn from an isotropic $k$-cluster Gaussian mixture:
\begin{align}
    \vec{x}_{\nu} \sim_{i.i.d.} \sum\limits_{c\in\mathcal{C}}p_{c}\mathcal{N}\left(\sqrt{\sfrac{\lambda}{s}}\vec{\mu}_{c}, \mat{I}_{d}\right), && \nu\in[n]
    \label{eq:main:sparsegmm}
\end{align}
\noindent where $p_{c}$ are the class probabilities, $\mathcal{C}$ is the index set ($|\mathcal{C}| = k$), $\vec{\mu}_{c}\in\mathbb{R}^{d}$ are the \emph{$s$-sparse} vectors representing the means of the clusters and $\lambda$ is a measure of the signal-to-noise ratio (SNR) for this model. In the following, we focus in the balanced case for which $p_{c}=\sfrac{1}{k}$. %We group together the cluster means component wise, i.e. considering the vector $\vec{v}_{i} =  (\mu_c^i)_{c\in\mathcal{C}} \in \mathbb{R}^k$,% and we assume that it is distributed according to a Gauss-Bernoulli distribution:
We select the subspace of relevant features thanks to the introduction of the vector $\vec{v}_{i}\in \mathbb{R}^k$. The projection of all the cluster means, on a given dimension $i \in [d]$, will be completely determined by a linear combination of the components of $\vec{v}_i$. We consider for this purpose a Gauss-Bernoulli distribution:
\begin{align}
\label{eq:gaussbernouilli}
    \vec{v}_{i}\sim_{i.i.d.} \rho \mathcal{N}(0,\mat{I}_{k})+(1-\rho)\delta_{0}
\end{align}
\noindent where $\rho \coloneqq \sfrac{s}{d}$ is the density of non-zero elements. For the convenience of the theoretical analysis that follows, it will be useful to work with a particular encoding of the class labels $\mathcal{C}$. For a given sample $\nu\in[n]$ belonging to the class $c\in\mathcal{C}$, define the following label indicator vector $\vec{u}_{c}^{\nu}\in\{-\frac{1}{k},\frac{k-1}{k}\}^{k}$:
\begin{align}
\label{eq:indicator}
\vec{u}_{c}^{\nu} &= \frac{1}{k} (-1,\dots,-1, \underbrace{k-1}_{c\text{-th element}},-1,\dots, -1) %\coloneqq \Tilde{\vec{e}}_c.
\end{align}
For a given draw $(\vec{x}_{\nu})_{\nu\in[n]}$, define the matrices $\mat{X}\in\mathbb{R}^{d\times n}$ with columns given by $\vec{x}_{\nu}$ and $\mat{U}\in\mathbb{R}^{n\times k}, \mat{V}\in\mathbb{R}^{d\times k}$ with rows given by $\vec{u}^{\nu}_{c}$ and $\vec{v}_i$. 

A crucial point that we shall exploit in this paper is that, with this notation, our model for subspace clustering can be written as a matrix factorization problem, where the data has been generated as:
\begin{align}
    \label{eq:model}
    \mat{X} = \sqrt{\frac{\lambda}{s}}\mat{V}\mat{U}^{\top} + \mat{W}
\end{align}
\noindent with $\mat{W}\in\mathbb{R}^{d\times n}$ a Gaussian matrix with elements $W_{i\nu}\sim\mathcal{N}(0,1)$. This formulation is completely equivalent to the one in eq.~\eqref{eq:main:sparsegmm}, by identifying the cluster means as given component-wise by: $\mu_c^{(i)} = \vec{v}_i^{\top} \vec{u}_c, \, i \in [d]$. The centroids will have, in expectation, only $s$ non-zero components; as discussed in the introduction, the $s$-sparse class means $\{\vec{\mu}_{c}\}_{c\in\mathcal{C}}$ can be thought of as low-dimensional features embedded in a higher-dimensional space $\mathbb{R}^{d}$, that have been corrupted by isotropic additive Gaussian noise with variance $\sim \lambda^{-1}$. Given a finite draw $(\vec{x}_{\nu})_{\nu\in[n]}$ generated from this model with class means $\mat{V}_{\star}$ and labels $\mat{U}_{\star}$, the goal of the statistician is to perform \emph{clustering}, i.e. to estimate $\mat{U}_{\star}$ from $\mat{X}$, which is equivalent to retrieving the class label for each sample in the data. However, note that there is a clear class symmetry in this problem: the labels can only be estimated up to a permutation $\pi\in\sym(\mathcal{C})$ reshuffling the columns of $\mat{U}_{\star}$. Taking this symmetry into account we can assess the performance of an estimator $\hat{\mat{U}}$ through the averaged symmetrized mean-squared error:
\begin{align}
    \label{eq:mse}
    \mse(\hat{\mat{U}}) = \underset{\pi\in\sym(\mathcal{C})}{\min}\frac{1}{n}\mathbb{E}||\pi(\hat{\mat{U}}) - \mat{U}_{\star}||_{F}
\end{align}
\noindent where $||\cdot||_{F}$ denotes the matrix Frobenius norm. In particular, we will be interested in characterizing reconstruction in the \emph{proportional high-dimensional limit} where the number of samples $n$, the ambient dimension $d$ and the sparsity level $s$ go to infinity $n,d,s\to\infty$ at fixed density $\rho = \sfrac{s}{d}$, sample complexity $\alpha = \sfrac{n}{d}$, number of clusters $k\geq 2$ and signal strength $\lambda>0$.

Note that the clustering problem above is closely related to the problem of estimating the class means / features $\mat{V}_{\star}$. Indeed, written as in eq.~\eqref{eq:model} the problem of estimating both the labels and centroids $(\mat{U}_{\star},\mat{V}_{\star})$ boils down to a low-rank matrix factorization problem. In this manuscript, we have chosen to frame the discussion in terms of the clustering, but all our results could be presented also in terms of the reconstruction of the class means. 

\begin{figure}
\centering
     \includegraphics[width=0.7\textwidth]{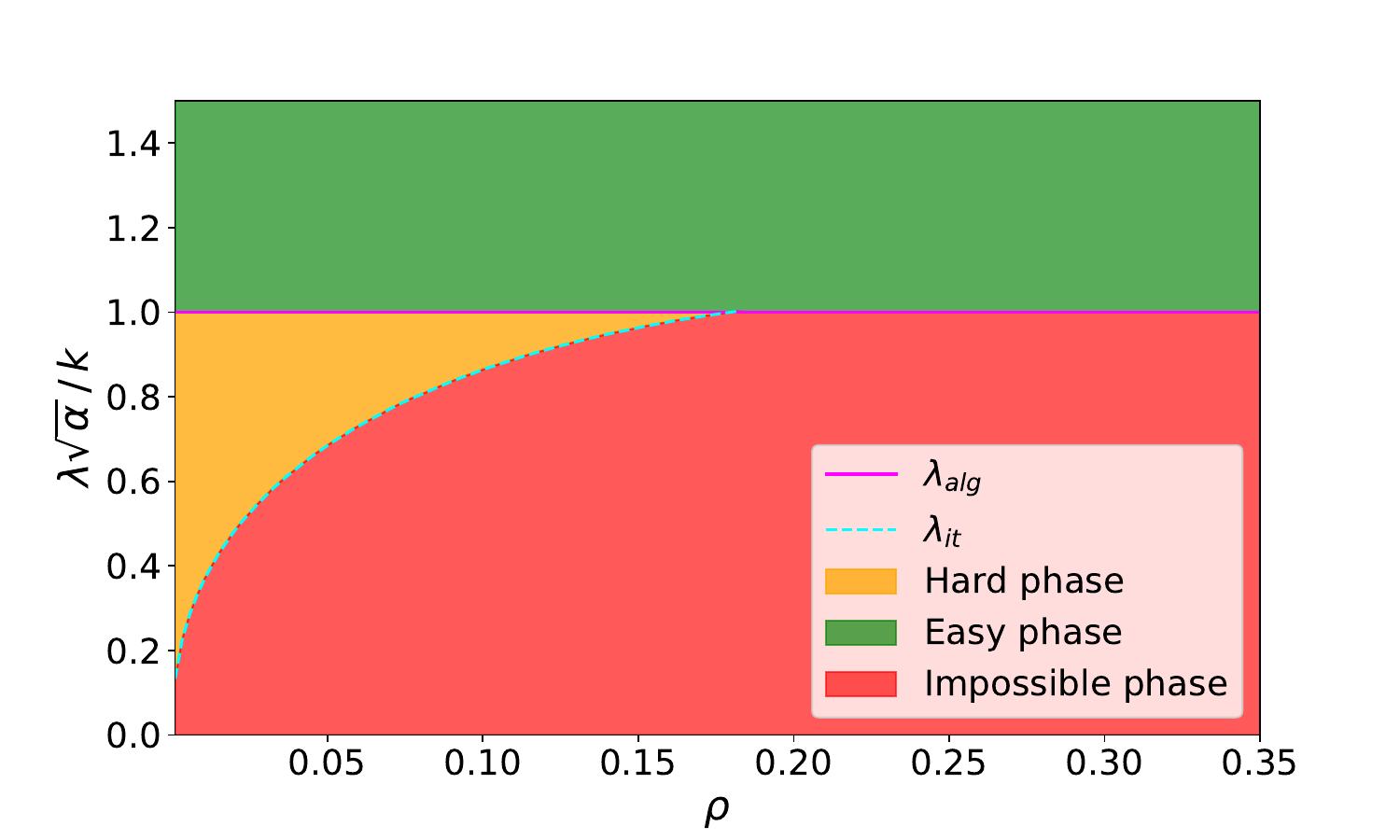}
     \vspace{-0.2cm}
      \caption{Phase diagram for the subspace clustering of two-clusters GMM at fixed $\alpha=2$.  We plot the SNR $\lambda$ as a function of $\rho$ and we rescale the y-axis by $\sfrac{\sqrt{\alpha}}{k}$. We colour different region of the figure according to the associated phase. The algorithmic threshold $\lambda_{\text{alg}}$ is the solid line in magenta while the information-theoretic threshold $\lambda_{\text{it}}$ is the dashed line in cyan. In the \emph{impossible} region, no method can perform better than a random guess. In the \emph{hard} region, a partial reconstruction of the signal is theoretically possible, but we do not know of any polynomial time algorithm that can do it. In the \emph{easy} region, however, AMP, can achieve positive correlation with the ground truth 
      (and actually achieves Bayes MMSE, except very close to the tri-critical point, see the discussion in App.~\ref{sec:app:potential})
      \vspace{-0.2cm}
      }
\label{fig:pd}
\end{figure}

In this manuscript we provide a sharp asymptotic characterization of when reconstruction, as measured by positive correlation with the ground truth, is possible in high-dimensions for this model, both \emph{statistically} and \emph{algorithmically}. In particular, our {\bf main contributions} are:
%\paragraph{Main contributions} 
\begin{itemize}[wide=1pt,noitemsep]
    \item We map the subspace clustering problem to an asymmetric matrix factorization problem. We compute the closed-form asymptotic formula for the performance of the Bayesian-optimal estimator in the high-dimensional limit, building on a statistical physics inspired approach \cite{PHD} that has been rigorously proven in this case \cite{Miolane2016,Miolane2017} . This allows us to provide a sharp threshold bellow which reconstruction of the features is \emph{statistically} impossible as a function of the parameters of the model.
    \item To estimate the algorithmic limitations of reconstruction, we analyse a tailored approximate message passing (AMP) algorithm \cite{deshpande2014information,fletcher2018iterative,PHD} approximating the posterior marginals in this problem, and derive the associated \emph{state evolution equations} characterising its asymptotic performance. Such algorithms are optimal among first order methods  \cite{Celentano2020, Celentano2021}) and therefore their reconstruction threshold provides a bound on the algorithmic complexity clustering in our model.
    \item The two results above allow us to paint a full picture of the \emph{statistical-to-algorithmic} trade-offs in high-dimensional subspace clustering for the sparse $k$-Gaussian mixture model, and in particular to identify an \emph{algorithmically hard} region of the sparsity level ($1-\rho$) vs. signal-to-noise ratio ($\lambda$) plane for which reconstruction is possible statistically but not algorithmically, see Fig.~\ref{fig:pd}. Further, we provide a detailed analysis in the high sparsity regime ($\rho \to 0^{+}$) of how the algorithmically hard region grows as we increase the number of clusters and the sparsity level. In particular, the  \emph{information theoretical} transition arises at $\lambda_{\text{it}} \approx 
    \sfrac{\sqrt{-k \rho \log{\rho}}}{\sqrt{\alpha }}$, and the \emph{algorithmic} one at 
    $\lambda_{\text{alg}} \ge \sfrac{k}{\sqrt{\alpha}}$.
    \item The analysis for AMP optimality relies on the finite $\rho$ assumption as $n,d \to \infty$. We thus also investigate the complementary case when the number of non-zero components is of the order $s  \lesssim {n}$ and indeed see that sparse principal component analysis (SPCA) and Diagonal thresholding \cite{JLU09} can perform better than random in this region. We find, however, that this requires $s \le \sqrt{n}$ (thus
$\rho = o(1)$). We rephrase our findings in terms of existing literature on the subspace clustering for two-classes Gaussian mixtures \cite{JIN,LOFFL}.
\end{itemize}
\paragraph{Related works:} Subspace clustering is a well-studied topic in classical statistics, with a wide range of methods used in practice, see \cite{REV2} for a review. Closer to this work is the theoretical line of work studying the limitations of clustering in high-dimensions. Baik, Ben Arous and Péché have shown that PCA for Gaussian mixture clustering fails to correlate with the mixture means below certain threshold known as the \emph{BBP transition} \cite{BBP}. For dense Gaussian means, the statistical and algorithmic limitations of clustering were analysed in different regimes of interest. Our approach to study subspace clustering relies on a mapping to a low-rank matrix factorization problem. Low-rank matrix factorization has been extensively studied in the literature, and its asymptotic Bayes-optimal performance was characterized in \cite{deshpande2014information,PHD,dia2016mutual,Miolane2016,Miolane2017,el2018detection,el2020fundamental}. The construction of AMP algorithms and the associated state evolutions for matrix factorization was done in \cite{tanaka13,deshpande2014information,PHD,fletcher2018iterative}. In this work, we leverage these general results on matrix factorization. The closest to our work is perhaps \cite{GMM_CL, PHD}, where the authors characterize the asymptotic reconstruction thresholds for the dense case ($\rho=1$) in the proportional limit where the number of samples and input dimensions diverge at a constant rate. Non-asymptotic results were also discussed in \cite{NDA} which considered a modification of Lloyd's algorithm \cite{LLOY} achieving minimax optimal rate and proving a computational lower bound. To the best of our knowledge, the case in which the means are sparse has only been analysed in the regime where the number of non-zero components is sub-extensive with respect to the input dimension. Lower bounds for the statistically optimal recovery threshold in this regime were given in \cite{Azizyan2013, VERZ}, while computational lower bounds are studied in \cite{FAN,bresler2019}. % {MODIFICA QUI}
%\cite{FAN} provides a computationally feasible minimax lower bound from the statistical query model \cite{Kearns1998}.% 
Clustering of two component Gaussian mixtures has been studied in order to shed light on comparison between statistical and algorithmical tractability in a sparse scenario in \cite{AZ14,JIN,LOFFL}. In particular, \cite{JIN} conjectured and \cite{LOFFL} proved algorithmic bounds for this problem. They claim that even below the BBP threshold they can build an algorithm  achieving exponentially small misclustering error, given that (up-to log factors) we have $s \lesssim \sqrt{n}$. We relate our findings to their work exploring the extreme sparsity regime in detail. 

%%%%%%%%%%%%%%%%%%%%%%%%%%%%%%%%%%%%%%%%%%%%
\section{Main theoretical results}
\label{sec:main:theory}
In this section, we introduce the two main technical results allowing us to characterize the limitations of  clustering reconstruction (both statistically and algorithmically) for the model introduced above in the proportional high-dimensional limit. 
\vspace{-0.2cm}
\paragraph{Statistical reconstruction: } First, note that up to the permutation symmetry, the estimator minimizing the averaged mean-squared error in eq.~\eqref{eq:mse} admits a closed-form solution given by the marginals of the posterior distribution:
\begin{align}
\label{eq:boestimator}
    \hat{\mat{U}}_{\bo} = \underset{\mat{U}\in\mathbb{R}^{n\times k}}{\argmin}~\mse(\mat{U}) = \mathbb{E}\left[\mat{U}|\mat{X}\right]
\end{align}
\noindent where the posterior distribution for the model defined in eq.~\eqref{eq:model} explicitly reads:
\begin{align}
    P(U|X) =\frac{1}{Z(X)}\prod_{\nu=1}^n P_{u}(\vec{u}_{\nu})\int_{\mathbb{R}^{k}}\prod\limits_{i=1}^{d}\left(\dd\mat{v}_{i}P_{v}(\vec{v}_i)\right)  \prod\limits_{\nu=1}^{n}\prod\limits_{i=1}^{d}e^{-\frac{1}{2}\left(X_{\nu i} - \sqrt{\frac{\lambda}{s}}\vec{u}_{\nu}^{\top}\vec{v}_i\right)^2}
    \label{eq:posterior}
\end{align}
\noindent and for convenience we defined the vectors $\vec{u}_{\nu}, \vec{v}_{i}\in\mathbb{R}^{k}$ which are the rows of $\mat{U},\mat{V}$, and with the prior distribution $P_{u}$ being the uniform distribution over the indicator vectors defined in eq.~\eqref{eq:indicator} and $P_{v}$ given by the Gauss-Bernouilli distribution defined in eq.~\eqref{eq:gaussbernouilli}. 

Although in principle it would be possible to compute the minimum mean-squared error (MMSE) given by the Bayes-optimal estimator in eq.~\eqref{eq:boestimator} by sampling from the posterior from eq.~\eqref{eq:posterior}, this is impractical when $d, n$ are large. For instance, simply computing the different integrals involved in the evidence $Z$ scale exponentially with the dimensions. The first  result consists precisely in a closed-form solution for the asymptotic performance of the Bayes-optimal estimator:
\begin{result}
\label{res:stat}
    In the proportional high-dimensional limit where $n,d,s\to\infty$ with fixed ratios $\rho = \sfrac{s}{d}$, $\alpha=\sfrac{n}{d}$ and fixed $\lambda, k$, the minimum mean-squared error for the reconstruction of $\mat{U}\in\mathbb{R}^{n\times k}$ is given by:
    \begin{align}
    \label{eq:res:mmse}
        \lim\limits_{n\to\infty}\mmse = \frac{k-1}{k} - \Tr{\mat{M}_u^{\star}}
    \end{align}
    \noindent where $\mat{M}^{\star}_{u}\in\mathbb{R}^{k\times k}$ is the solution of the following minimization problem:
    \begin{align}
    \label{eq:res:argmin}
        \mat{M}_{u}^{\star} = \underset{\mat{M}_{u}\in\mathbb{R}^{k\times k}}{\argmin}~\left\{\max_{\mat{M}_{v}\in\mathbb{R}^{k\times k}}~ \Phi_{\rs}\left(\mat{M}_{u},\mat{M}_{v}\right)\right\}.
    \end{align}
    with:
    \begin{align}
        \Phi_{\rs}(\mat{M}_u,\mat{M}_v) = &\frac{\alpha \lambda \Tr{\mat{M}_u \mat{M}_v}}{2\rho} - \EX _{\vec{v}_*,\vec{w}}\left[\log{Z_{v}\left(\frac{\alpha \lambda \mat{M}_u}{\rho} , \frac{\alpha \lambda \mat{M}_u \vec{u}_*}{\rho} + \sqrt{\frac{\alpha \lambda \mat{M}_u}{\rho}} \vec{w}\right)}\right] \notag\\ 
        &- \alpha \EX _{\vec{u}_*,\vec{w}}\left[\log{Z_{u}\left( \frac{\lambda \mat{M}_v}{\rho},  \frac{\lambda \mat{M}_v \vec{v}_*}{\rho} + \sqrt{\frac{\lambda \mat{M}_v}{\rho}} \vec{w}\right)}\right]
        \label{eq:app:phi}
    \end{align}
    \noindent where we introduced $\vec{u}_* \sim P_u, \vec{v}_* \sim P_v, \vec{w} \sim \mathcal{N}(0,\mat{I}_k)$, and we defined $Z_{u/v}$ as:
    \begin{align}
    Z_u(\mat{A},\vec{b}) &= \frac{1}{k}\sum_{c\in\mathcal{C}} \exp{\left(\vec{b}^{\top}\vec{u}_c - \frac{1}{2}\vec{u}_c^{\top}\mat{A}\vec{u}_{c}\right)} \label{eq:main:subspace_Zs1} \\
    Z_v(\mat{A},\vec{b}) &= 1-\rho + \rho \exp \left( \frac{\vec{b}^{\top}(\mat{I}_k+\mat{A})^{-1}\vec{b}}{2} \right)\sqrt{\det \left( \mat{I}_k + \mat{A} \right)^{-1}}  
    \label{eq:main:subspace_Zs2}
\end{align}
\end{result}
Result \ref{res:stat} follows from our mapping of the subspace clustering problem introduced in Sec.~\ref{sec:main:model} to a low-rank matrix factorization form eq.~\eqref{eq:model}. Indeed, this mapping allows us to leverage a closed-form formula characterizing the asymptotic MMSE for low-rank matrix estimation with generic priors that was derived heuristically \cite{GMM_CL, PHD} using the replica method from Statistical Physics and was rigorously proven in a series of works \cite{deshpande2014information,dia2016mutual,Miolane2016, Miolane2017} in the context of subspace clustering. Our contribution resides in making this connection and drawing the consequences for the subspace clustering problem, a non-trivial endeavour given the complexity of the resulting formulas.
%Our contribution consists of mapping the subspace clustering problem discussed in Sec.~\ref{sec:main:model} to a low-rank matrix factorization form eq.~\eqref{eq:model}, allowing us to leverage the formalism from \cite{GMM_CL, PHD} to provide a detailed characterization of the limits of clustering. 
Note that the minimization problem eq.~\eqref{eq:res:argmin} is fundamentally different from the one in eq.~\eqref{eq:boestimator}. Indeed, the first involves only low-dimensional variables, and can be easily solved in a computer, while the latter is a high-dimensional problem which is computationally intractable for large $d,n$. The other parameter $\mat{M}_{v}^{\star}\in\mathbb{R}^{k\times k}$ solving eq.~\eqref{eq:res:argmin} can be used to characterize the MMSE reconstruction error on $\mat{V}$.
\paragraph{Algorithmic reconstruction: } While result \ref{res:stat} allow us to sharply characterize when clustering is statistically possible, it does not provide us a practical way to estimate the true class labels $\mat{U}_{\star}\in\mathbb{R}^{n\times k}$ from the data $\mat{X}\in\mathbb{R}^{d\times n}$. In order to provide a bound in the algorithmic complexity clustering, we consider an \emph{approximate message passing} (AMP) algorithm for our problem. Message passing algorithms are a class of first order algorithms (scaling as $O(nd)$, the dimensions of the input matrix $\mat{X}$) designed to approximate the marginals of a target posterior distribution, and which have two very important features. First, for a large class of random problems (such as the clustering problem studied here) AMP provides the best known first order method in terms of estimation performance \cite{Donoho2009, Krzakala2012, Barbier2019, Aubin2021, Aubin2020}, and has been rigorously proven to be the optimal for certain problems \cite{Celentano2020, Celentano2021}. Second, the asymptotic performance of AMP can be tracked by a set of low-dimensional \emph{state evolution equations} \cite{SE_PRF}, meaning that its reconstruction performance can be sharply computed without having to run a high-dimensional instance of the problem. For low-rank matrix factorization problems an associated AMP algorithm can be derived \cite{NIPS2013_5b69b9cb,tanaka13,deshpande2014information,fletcher2018iterative,Montanari2021}. Therefore, yet again we leverage the mapping of subspace clustering to a low-rank matrix factorization problem to derive the associated AMP algorithm \ref{alg:lowramp} with denoising functions $\eta_{v}, \eta_{u}$:
\begin{align}
    \eta_u(\mat{A},\vec{b}) &= \frac{1}{\sum_{c=1}^k \exp{\left(\vec{b}^{\top}\vec{u}_c - \frac{\vec{u}_c^{\top}\mat{A}\vec{u}_c}{2}\right)}} \sum_{c=1}^k \vec{u}_c \exp{\left(\vec{b}^{\top}\vec{u}_c - \frac{\vec{u}_c^{\top}\mat{A}\vec{u}_c}{2}\right)} 
    \label{eq:main:denoiser_u}\\
    \eta_v(\mat{A},\vec{b}) &= \frac{(\mat{I}_k+A)^{-1}\vec{b}}{\rho + (1-\rho)\sqrt{\det{\left(\mat{I}_k+\mat{A}\right)}}\exp{\left({\frac{-\vec{b}^{\top} (\mat{I}_k+\mat{A})^{-1}   \vec{b}}{2}}\right)}}
    \label{eq:main:denoiser_v}
\end{align}
As mentioned above, one of they key features of Algorithm \ref{alg:lowramp} is that its asymptotic performance can be tracked exactly by a set of low-dimensional equations.
%\florent{In which way is this our result? This is litteratly the State Evolution use in Thibault, proven in various work, for instance Deshpande, Fletcher, etc etc....}

\begin{result}
\label{res:se}
In the proportional high-dimensional limit where $n,d,s\to\infty$ with fixed ratios $\rho = \sfrac{s}{d}$, $\alpha=\sfrac{n}{d}$ and fixed $\lambda, k$, the correlation between the ground truth $(\mat{U}_{\star}, \mat{V}_{\star})$ and the AMP estimators $(\hat{\mat{U}}_{\amp}^{t}, \hat{\mat{V}}_{\amp}^{t})$ at iterate $t$,
\begin{align}
    \mat{M}^{t}_{u} = \frac{1}{n}\mat{U}^{\top}_{\star}\hat{\mat{U}}^{t}_{\amp}, && \mat{M}^{t}_{v} = \frac{1}{n}\mat{V}^{\top}_{\star}\hat{\mat{V}}^{t}_{\amp}
\end{align}
satisfy the following \emph{state evolution equations}:
 \begin{align}
     \mat{M}_u^{t+1} &= \EX_{\vec{u}_*\sim P_{u},\vec{\xi}\sim \mathcal{N}(\vec{0}_{k},\mat{I}_{k}})\left[\eta_u\left(\frac{\alpha \lambda \mat{M}_u}{\rho} , \frac{\alpha \lambda \mat{M}_u \vec{u}_*}{\rho} + \sqrt{\frac{\alpha \lambda \mat{M}_u}{\rho}} \vec{w}\right)\vec{u}_{*}^{\top}\right]  \\
     \mat{M}_v^{t+1} &= \EX_{\vec{v}_*\sim P_{v},\vec{\xi}\sim \mathcal{N}(\vec{0}_{k},\mat{I}_{k}})\left[\eta_v\left( \frac{\lambda \mat{M}_v}{\rho},  \frac{\lambda \mat{M}_v \vec{v}_*}{\rho} + \sqrt{\frac{\lambda \mat{M}_v}{\rho}} \vec{w}\right)\vec{v}_*^{\top}\right] 
     \label{eq:se_eq}
 \end{align}

Moreover, the asymptotic performance of is simply given by: %\florent{Why ?? (19)?} $\hat{\mat{U}}^{t}_{\amp}$ 
 \begin{align}
     \lim\limits_{n\to\infty} \mse(\hat{\mat{U}}_{\amp}^{t}) = \frac{k-1}{k} - \Tr{\mat{M}_u^t}
 \end{align}
 \end{result}
Result \ref{res:se} is a consequence from the general theory connecting AMP algorithms to their state evolution (SE) \cite{SE_PRF}. In the context of low-rank matrix factorization, a derivation of the state evolution equations above from Algorithm \ref{alg:lowramp} were first provided for the rank-one case in \cite{deshpande2014information,fletcher2018iterative} and were extended to general rank and denoising functions in \cite{PHD,GMM_CL}. A crucial observation is that Result \ref{res:se} is intimately related to Result \ref{res:stat}.
Indeed, the state evolution equations \eqref{eq:se_eq} coincide exactly with running gradient descent on the potential defined in eq.~\eqref{eq:app:phi}.
While the performance of the statistically optimal estimator $\hat{\mat{U}}_{\bo}$ is given by the fixed point with the minimal value of the potential $\Phi_{\rs}$, the performance of the AMP estimator $\hat{\mat{U}}_{\amp}$ is described by the closest minima to the initialization $\mat{M}_{u}^{t=0}$. Therefore, studying both the statistical and algorithmic limitations of clustering in high-dimensions boils down to the study of the minima of the potential $\Phi_{\rs}$, see App.~\ref{sec:app:potential} for further details. 
The identification of a subspace clustering problem with a matrix factorization one allows us to leverage the rich literature from this field. However, analyzing these formulas is far from trivial and consitutes a considerable technical challenge.
The major difficulty is to find a suitable parametrization of the overlap
matrices that reduces the number of parameters to be tracked, we discuss this in detail in Sec.~\ref{sec:main:k2}. Moreover, we find the scaling ansatz in the high-sparsity limit that closes the equations on amenable quantities in order to find analytically the statistical-to-computational gap for the large-rank setting, see Sec.~\ref{sec:main:small}. 
However, we stress that we did not take all the necessary precautions to claim full rigor.  e.g. prove formally that the minimum is unique (although we checked all these both analytically and numerically).
\looseness=-1
\begin{algorithm}[bt]
   \otherlabel{alg:lowramp}{1}
   \caption{low-rAMP}
\begin{algorithmic}
   \STATE {\bfseries Input:} Data $\mat{X}\in\mathbb{R}^{d\times n}$
   \STATE Initialize $\hat{\vec{v}}_{i}^{t=0}, \hat{\vec{u}}_{\nu}^{t=0}\sim\mathcal{N}(\vec{0}_{k},\epsilon\mat{I}_{k})$, $\hat{\sigma}_{u, \nu}^{t=0} = \mat{0}_{k\times k}$, $\hat{\sigma}_{v,i}^{t=0} = \mat{0}_{k\times k}$.
  
   \FOR{$t\leq t_{\text{max}}$}
   \STATE $\mat{A}_{u}^{t} = \frac{\lambda}{s} \left(\hat{\mat{U}}^t\right)^{\top} \hat{\mat{U}}, \qquad A_{v}^{t} = \frac{\lambda}{s}\left(\hat{\mat{V}}^t\right)^{\top} \hat{\mat{V}}$ 
   \STATE $\mat{B}_{v}^{t} = \sqrt{\frac{\lambda}{s}}X\hat{\mat{U}}^t-\frac{\lambda}{s}\sum\limits_{\nu=1}^{n}\sigma^{t}_{u, \nu}\hat{\mat{V}}^{t-1}$, \quad  $\mat{B}_{u}^{t} = \sqrt{\frac{\lambda}{s}}X^{\top}V-\frac{\lambda}{s}\sum\limits_{i=1}^{d}\sigma^{t}_{v, i}\hat{\mat{U}}^{t-1}$
   \STATE Take $\{\vec{b}^t_{v,i} \in \mathbb{R}^k\}_{i=1}^d, \{\vec{b}^t_{u,\nu}\in \mathbb{R}^k\}_{\nu = 1}^n$ rows of $\mat{B}^t_{v},\mat{B}^t_u$
    \STATE $\hat{\vec{v}}_{i}^{t+1} = \eta_{v}(\mat{A}_{v}^{t}, \vec{b}_{v,i}^{t})$, \qquad $\hat{\vec{u}}_{\nu}^{t+1} = \eta_{u}(\mat{A}_{u}^{t}, \vec{b}_{u,\nu}^{t})$
   \STATE $\hat{\sigma}_{v,i}^{t+1} = \partial_{\vec{b}}\eta_{v}(\mat{A}_{v}^{t}, \vec{b}_{v,i}^{t})$, \qquad  $\hat{\sigma}_{u,\nu}^{t+1} = \partial_{\vec{b}}\eta_{u}(\mat{A}_{u}^{t}, \vec{b}_{v,\nu}^{t})$

   \STATE Here $\hat{\mat{U}}^t \in \mathbb{R}^{n\times k},\hat{\mat{V}}^t \in \mathbb{R}^{d\times k},\mat{B}_u^t \in \mathbb{R}^{n\times k}, \mat{B}_v^t \in \mathbb{R}^{d\times k},\mat{A}_u^t \in \mathbb{R}^{k\times k}, \mat{A}_v^t \in \mathbb{R}^{k\times k}$
   \ENDFOR
   \STATE {\bfseries Return:} Estimators $\hat{\vec{v}}_{\amp,i}, \hat{\vec{u}}_{\amp,\nu}\in\mathbb{R}^{k}, \hat{\sigma}_{u, \nu}, \hat{\sigma}_{v, i}\in\mathbb{R}^{k\times k}$
\end{algorithmic}

\end{algorithm}

%%%%%%%%%%%%%%%%%%%%%%%%%%%%%%%%%%%%%%%%%%%%
\section{Reconstruction limits for sparse clustering}
\label{sec:main:thresh}
\begin{figure}
\centering
     \includegraphics[width=0.95\textwidth]{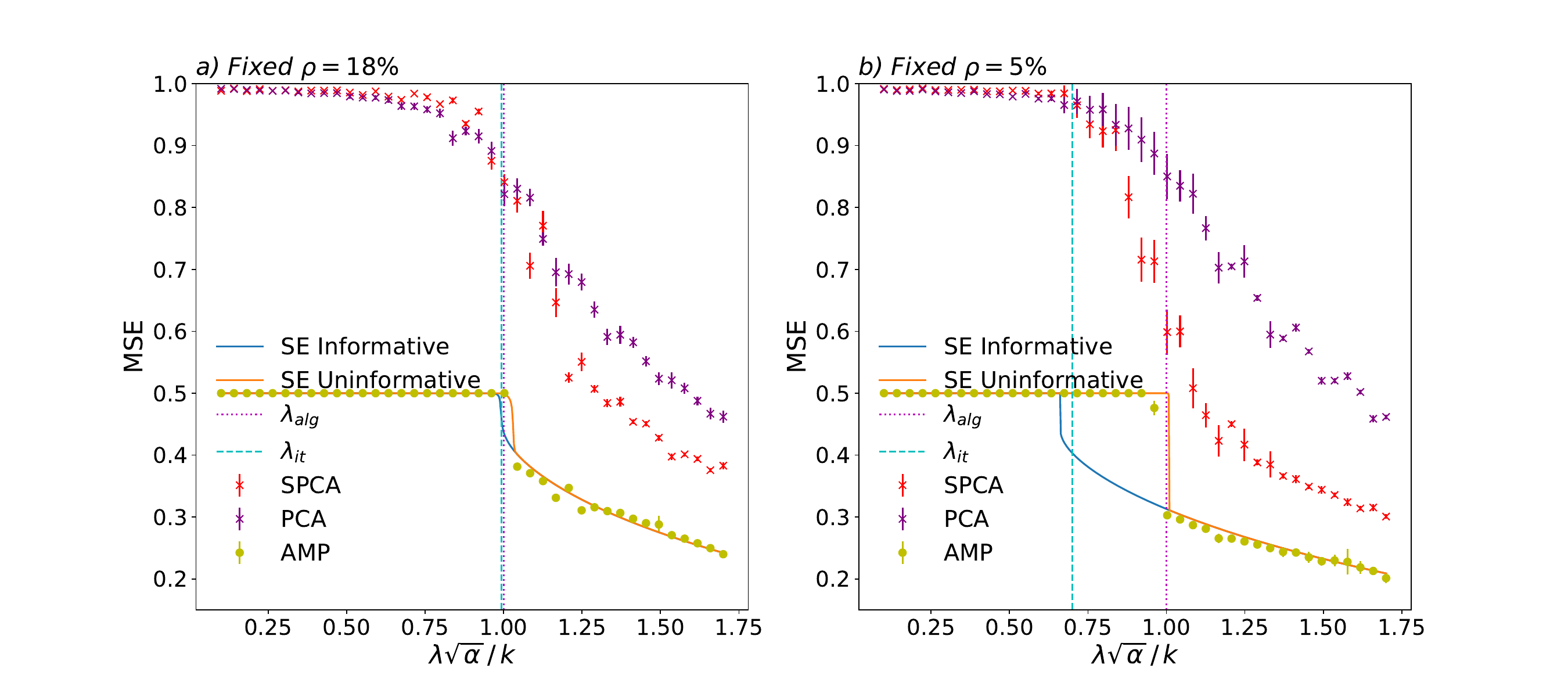} 
          \vspace{-0.2cm}
      \caption{We compare the performance for clustering of two-classes GMM, as measured by the MSE of AMP, SPCA, PCA and SE informed and uninformed. We plot the MSE as a function of the SNR $\lambda$ and we rescale the x-axis by $\sfrac{\sqrt{\alpha}}{k}$. For each algorithm considered the error bars are built using the standard deviation over fifty runs with parameters $(n=8000, d=4000)$, i.e. $\alpha=2$. We plot in vertical line the theoretical values for the Information-Theoretic threshold $\lambda_{\text{it}}$ (dashed cyan line), and the  algorithmic threshold $\lambda_{\text{alg}}$ (dotted line in magenta). The theoretical values coincide with the experimental one. The SE with uninformed initialization follows AMP as expected. \emph{Left}: The sparsity is fixed with parameter $\rho= 18 \%$. Both SPCA and PCA have a worse performance with respect to AMP and in this sparsity regime we have only a marginal advantage by using SPCA with respect to PCA.  \emph{Right}: The sparsity is fixed with parameter $\rho= 5 \%$. Increasing the sparsity level the width of the algorithmically hard phase becomes bigger and SPCA performs clearly better than PCA.  %{\color{blue}{Luca: I didn't put Diagonal Thresholding in the plot for now but you may see how it would look like in Fig.~\ref{fig:mses+}}}
      \vspace{-0.4cm}
      }
\label{fig:mses}
\end{figure}

These results allow us to paint a full picture for when sparse subspace clustering is possible in the model defined in Sec.~\ref{sec:main:model} as a function of the quantity of data $\alpha$, the sparsity  $1\!-\!\rho$, the number of clusters $k$ and the SNR $\lambda$. Figure \ref{fig:pd} summarizes the different reconstruction regimes in the $(\rho, \lambda)$ plane for a two-clusters problem at fixed sample complexity $\alpha\!=\!2$, also known as a \emph{phase diagram}. 
%In order to somplifying computationally the problem, we map the two-cluster problem into an equivalent rank-one matrix factorization problem. The $k=1$ version of model presented in Sec.~\ref{sec:main:model} is ill defined and we detail in App.~\ref{sec:app:pd} the correct mapping.%
Moreover the general considerations on the  reconstruction limits for sparse clustering, given by analyzing the two-clusters problem, are easily generalizable to the general mixture case and not restricted to that particular model, see Sec.~\ref{sec:main:k2}. For a fixed sparsity $1\!-\!\rho$, we identify the following regions in Fig.~\ref{fig:pd}:\looseness=-1
\vspace{-0.22cm}
\begin{description}[wide=1pt]
    \item[Impossible phase:] There is not enough information in the data matrix $X$ handled to the statistician to assign cluster membership better than chance for \textit{any algorithm}. The Bayes-optimal MMSE is not better than a random guess. Clustering (reconstruction of $\mat{U}_{\star}$ better than chance) is impossible.
    \item[Hard phase: ] The MMSE is non-trivial, and clustering is statistically possible to some extent, but the best known polynomial time algorithm, AMP, fails to correlate better than chance with the true cluster assignment $\mat{U}_{\star}$. Any polynomial-time algorithm  is conjectured to fail in this region. 
    \item[Easy phase: ] In the easy phase, not only clustering is statistically possible, but AMP is able to achieve positive correlation with $\mat{U}_{\star}$. One can also investigate when AMP achieves the Bayes-optimal MMSE (instead of just positive correlation) leading to the same transition (except from a subtle correction very close to the tri-critical point, see the discussion in App.~\ref{sec:app:potential}).
\end{description}
\vspace{-0.22cm}
In Fig.~\ref{fig:mses}, we investigate the different phases presented above by varying the sparsity level $1-\rho$ and SNR $\lambda$ at fixed $\alpha$. We compare the performance of AMP with popular spectral algorithms: Principal Component Analysis (PCA) and Sparse Principal Component Analysis (SPCA).  We can initialize AMP and the SE equations in two different ways: we call \emph{uninformative initialization} a choice for the first iterates of AMP and SE which assumes no knowledge on the ground truth values; %This would correspond to $M_u^{t=0} \simeq 0$ in the SE structure and to a small-$\varepsilon$ parameter in the initialization of Algorithm~\ref{alg:lowramp}.%
conversely with \emph{informative initialization} we consider that the statistician has some prior knowledge on the the ground truth signal. 
%This would translate into the choice $M_u^{t=0} \simeq 1, M_u^{t=0}\simeq $ in SE language and AMP would be initialized at the ground truth values.
Note that in a real-life scenario the statistician does not have access to the ground truth. Yet, as a theoretical tool the informative initialization provides important information about the algorithmically hard phase, see App~\ref{sec:app:potential} for a discussion. The initialization strategy we considered in the uninformed case for Algorithm~\ref{alg:lowramp} is not the only possible choice, there are smarter ways of initialising which can lead to a considerable improvement without explicitly assuming any information about the signal, e.g. spectral initialization \cite{mondelli21}. %\florent{the following line must be the least efficient way to use space in a paper i have ever seen. Joke asside, the uninformative is ambigous... should probably cite https://arxiv.org/abs/2008.03326 at this point.} 
Looking at Fig.~\ref{fig:mses} we see that SE with uninformed initialization tracks AMP. Morover we note that increasing the sparsity level, i.e. decreasing $\rho$, the problem becomes algorithmically harder. This is reflected in a discontinous jump in the MSE at $\lambda_{\alg}$ which becomes larger. We observe along the same lines a neat advantage by imposing the sparsity constraint in the spectral algorithm (SPCA) with respect to vanilla one (PCA) as the sparsity grows. We discuss the details on the numerical simulations in App.~\ref{sec:app:numerics} and the code is available at \href{https://github.com/lucpoisson/SubspaceClustering}{https://github.com/lucpoisson/SubspaceClustering}. 

%%%%%%%%%%%%%%%%%%%%%%%%%%%%%%%%%%%%%%%%%%%%
\section{Stability analysis and algorithmic threshold}
\label{sec:main:k2}
% \florent{This section is full of new contribution, this should be clearer. }{\color{blue}{Luca: Is this going in a better direction?}}
% We study now the phase transitions from a theoretical standpoint. As discussed in the end of  Sec.~\ref{sec:main:theory}, we can derive a tailored AMP for sparse GMM clustering by manipulating the expressions appearing in eq.~\eqref{eq:main:subspace_Zs}
% \vspace{-0.1cm}
In this section we provide a detailed analytical derivation of the threshold $\lambda_{\alg}$ characterizing the algorithmic reconstruction as a function of the number of clusters and the sparsity.
% Unfortunately its SE equations do not admit an analytical closed formula for the iterates $(\mat{M}_u^t,\mat{M}_v^t)$.
% In order to pursue an analytical approach to the analysis of the phase transitions we exploit the exchange symmetry between different clusters. 
First, note due to the permutation symmetry of the clusters the overlap matrices admit the following parametrization:
\begin{align}
\mat{M}_u^t= \frac{m_u^t}{k} \mat{I}_k - \frac{m_u^t}{k^2} \mat{J}_k && 
\mat{M}_v^t = m_v^t \mat{I}_k - \frac{m_v^t}{k} \mat{J}_k
\label{eq:scalar_param}
\end{align}
\noindent where $\mat{J}_k$ is the $k\times k$-matrix with all elements equal to one. This parametrization is preserved under the SE iterations.
Therefore, inserting it in the SE equations yield equations for $(m_u^t,m_v^t)$:
\begin{align}
m_u^{t+1}=  f_u^{(k)}(\sfrac{\lambda m_v^t}{\rho})  && 
m_v^{t+1}=  f_v^{(k)}(\sfrac{\alpha \lambda m_u^t}{\rho})
\label{eq:se_simple}
\end{align}
\noindent where we introduced the following update functions:
\begin{align}
 f_u^{(k)}(z) &\coloneqq \frac{k}{k-1} \EX_{\vec{\omega}\sim \mathcal{N}(\vec{0}_{k},\mat{I}_{k}}) \left [ \frac{e^{z +w_1\sqrt{z}}}{e^{\lambda m_v^t +w_1\sqrt{z}} + \sum_{l=2}^k e^{w_l\sqrt{z}}}\right] - \frac{1}{k-1}  \\
 f_v^{(k)}(z) &\coloneqq   \frac{\rho^2z}{k(k+z)}\int_{0}^{+ \infty}\frac{S_{k-1}}{(2 \pi)^{\frac{k}{2}}}  \frac{\xi^{k+1}e^{-\sfrac{\xi^2}{2}}}{\rho + (1-\rho)(\frac{k+z}{k})^{\frac{k}{2}}e^{-\frac{\xi^2 z}{2k}}} \, \dd\xi
\end{align}
\noindent where $S_{k-1}$ is the surface of the $k$-dimensional unitary hypersphere. Recall that $(m_u^t,m_v^t) \! \in \! [0,1]^2$ fully characterize the reconstruction performance, with $(m_u^{\infty},m_v^{\infty}) \!=\! (0,0)$ corresponding to the performance of a random guess. Conversely, $(m_u^{\infty},m_v^{\infty}) \!=\! (1,\rho)$ corresponds to perfect reconstruction of the cluster membership and the sparse cluster means. One can check that the \emph{trivial fixed point} $(m_u,m_v)\!=\!(0,0)$ is always a fixed point of the SE equations. Moreover, note that its stability is crucially connected to the algorithmic threshold $\lambda_{\alg}$. Indeed, if the trivial fixed point $(m_u,m_v)\!=\!(0,0)$ is stable, AMP with an uninformed initialization will always converge to this point - and therefore achieve random guessing performance. To study its stability, we expand the update functions around $(m_u,m_v)\!=\!(0,0)$ up to the second order. Expressing everything in terms of $m_u$, we obtain:\looseness=-1
\begin{align}
m_u^{t+1} = F_{\text{se}}\left(m_u^t\right) = \frac{\lambda^2 \alpha}{k^2}m_u^t + \left (\frac{(k-4) \alpha^2 \lambda^4}{2k^4} - \frac{\lambda^3\alpha^2}{k^3}\right)\left(m_u^t\right)^2  + o\left((m_u^t)^2\right) 
\label{eq:main:pert_exp}
\end{align}
This immediately tells us that the trivial fixed point becomes unstable at the \textit{algorithmic threshold} $\lambda_{\text{alg}} = \sfrac{k}{\sqrt{\alpha}}$. This transition is a well-known result in random matrix theory and goes under the name of BBP transition \cite{BBP}. Despite the fact that we expand around the trivial fixed point, the perturbative method also provides information also about the Bayes-optimal performance, thanks to the general properties that the phase diagrams in Bayes-optimal inference problems must respect \cite{BAY_PT}. Indeed, a sufficient criterion for the presence of an algorithmically hard phase requires at the algorithmic threshold: $F^{\prime \prime}_{\text{se}}(0) > 0 $. The local study predicts that the phase diagram will present an algorithmically hard region for $k > k_{\text{hard}} = 4 + 2\sqrt{\alpha}$. The criterion nevertheless is not necessary in this setting, as we immediately see from the phase diagram for the two-components GMM in Fig.~\ref{fig:pd}: we clearly observe an algorithmically hard region for high-sparsity although $k<k_{\text{hard}}$. In fact, the \textit{information-theoretic threshold} $\lambda_{\text{it}}$, defined as the value of the SNR at which the problem becomes statistically possible, cannot be found simply thanks to the expansion around the trivial fixed point. Indeed the exact computation of $\lambda_{\text{it}}$ requires evaluating the potential function $\Phi$ in eq.~\eqref{eq:app:phi}: one needs to find the threshold value of the SNR at which the informative fixed point has a lower value of the free energy with respect to the uninformative fixed point. The technical discussion on the calculation of $\lambda_{\text{it}}$, which we see plotted in Fig.~\ref{fig:pd} for the two-classes case, is given in App.~\ref{sec:app:pd}, while a general overview on the different thresholds is given in App.~\ref{sec:app:potential}. It is interesting to stress that the linearization of the SE equation around the trivial fixed point in eq.~\eqref{eq:main:pert_exp} coincide to study the gradient of the potential $\Phi_{\rs}$ in eq.~\eqref{eq:app:phi} around that point, and hence the stability of the trivial fixed point is determined by the potential landscape. This connection is at the roots of the conjectured link between the hardness of an inference task and AMP: the statistical-to-computational gap opens up as the fixed point with minimal value of the potential in eq.~\eqref{eq:app:phi}, describing the performance of the Bayes-optimal estimator, is attained far from a locally stable region around the uninformative fixed point, where the AMP algorithm gets stuck.

%%%%%%%%%%%%%%%%%%%%%%%%%%%%%%%%%%%%%%%%%%%%
\section{Large sparsity regime}
\label{sec:main:small}
We characterize in this section the scaling of the thresholds in the very sparse (and most interesting) regime when $\rho \to 0^{+}$. We highlight the main passages of the computation and focus on the principal results, see Appendix~\ref{sec:app:large_spars} for a  detailed analysis. 
The starting point is the following change of variables:
\begin{align}
    m_u = \tilde{m}_u \sqrt{\frac{-\rho  \log{\rho}}{\alpha}} &&
    m_v = \tilde{m}_v \rho  && 
    \lambda = C(k) k \sqrt{\frac{-\rho \log{\rho}}{\alpha  }}
    \label{eq:ansatz}
\end{align}
Inserting these expressions into eq.~\eqref{eq:se_simple}, we obtain simplified SE equations for the rescaled overlaps $(\tilde{m}_u,\tilde{m}_v)$ without any residual dependence on $(\rho,\alpha)$: 
\begin{align}
    \Tilde{m}_u = C(k) \tilde{m}_v && \tilde{m}_v = T_k\left(C(k) \Tilde{m}_u\right)
    \label{eq:main:
    small_rho_se}
\end{align}
\noindent where we introduced the auxiliary function $T_k(\cdot)$ defined as: $$T_k(z) = \int_{0}^{+\infty} \frac{S_{k-1}}{k (2 \pi)^{\sfrac{k}{2}}}\xi^{k+1}e^{-\sfrac{\xi^2}{2}} \Theta \left(\frac{z\xi^2}{2} - 1 \right) \, d\xi$$  
\noindent where $\Theta(\cdot)$ is the Heaviside theta function. By considering the large $k$ expansion of $T_k(\cdot)$, we can derive the scaling of the IT threshold with $(k,\rho,\alpha)$, see Appendix \ref{sec:app:large_spars} for more details. Putting together with our previous result for $\lambda_{\alg}$ from Sec.~\ref{sec:main:k2}, we obtain the following fundamental result:
\begin{align}
    \lambda_{\text{it}} \approx \sqrt{\frac{-k \rho \log{\rho}}{\alpha}} &&
    \lambda_{\text{alg}} =  \frac{k}{\sqrt{\alpha}}.
\end{align}
These equations completely characterize the behavior at large rank \& small (but finite) sparsity. The  statistical-to-computational gap, where AMP is not able to  exploit  the information on the sparse nature of the cluster means, grows with both the sparsity and the rank. In App.~\ref{sec:app:large_spars}, we further show that the large rank expression for the IT threshold $\lambda_{\text{it}}$ is accurate already at moderate $k \approx 10$, see Fig.~\ref{fig:C_trans}.\looseness=-1
\begin{figure}[t]
\centering
     \includegraphics[width=0.95 \textwidth]{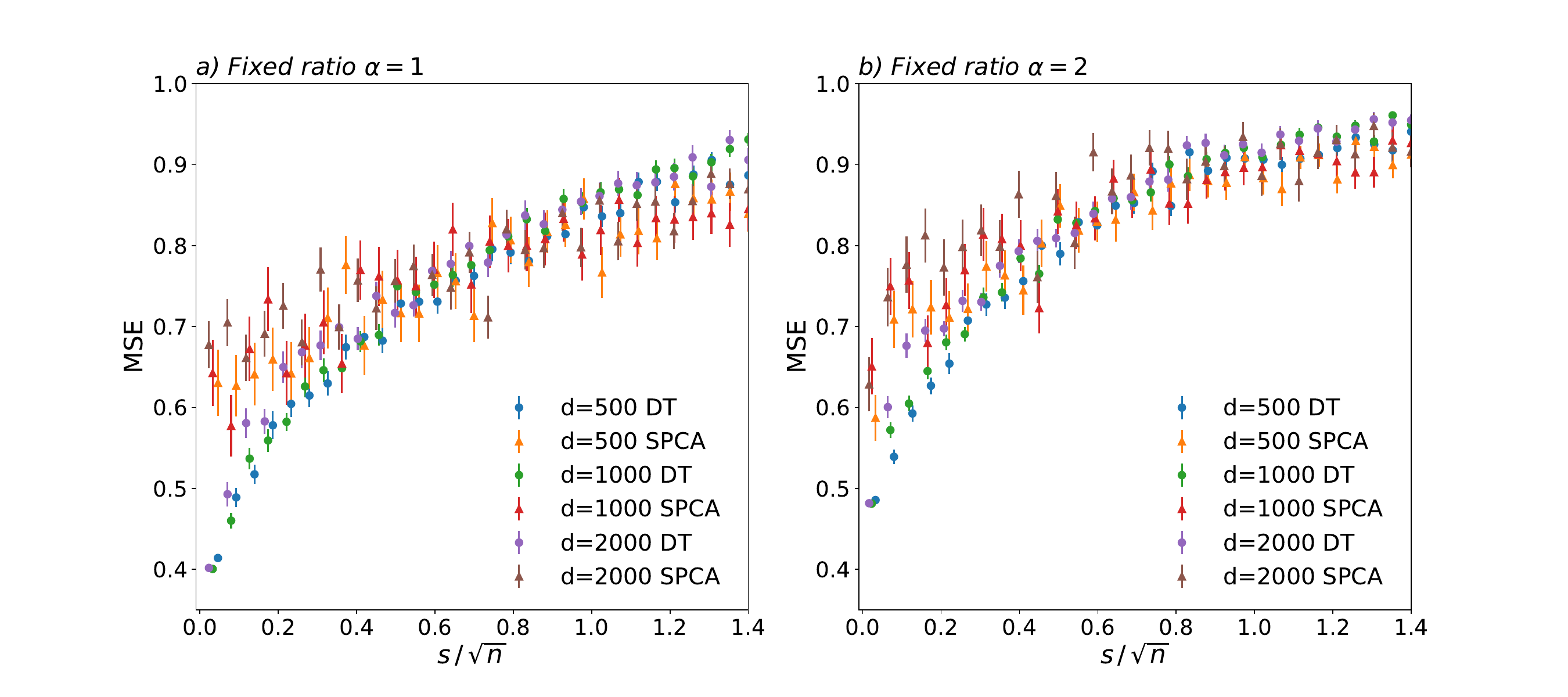} 
     \vspace{-0.3cm}
      \caption{We compare the performance of diagonal thresholding (DT) and Sparse PCA (SPCA), as measured by the MSE, for clustering of two-classes GMM for two different parameter $\alpha$. We plot the MSE vs the number of non-zero component $s$ and we rescale the x-axis by the square-root of the number of samples. The left plot is done at $\alpha=1$ while the right one is for $\alpha=2$.The SNR is tuned such that we are always under the BBP threshold.  With this choice we always work in the sub-extensive sparsity regime $\rho = o(1)$ and we can verify numerically what has been claimed in the literature \cite{JIN,LOFFL}: efficient algorithms in the high dimensional limit need a number of non-zero component (up-to log factors) $s 	\lesssim \sqrt{n}$ in order to beat random guessing under the BBP threshold.\vspace{-0.2cm}}
\label{fig:dtr}
\end{figure}
It is interesting at this point to reframe our results on the high sparsity regime in the context of the existing literature. The results for the IT transition for the detection of two-classes sparse GMM were discussed in \cite{FAN,VERZ}, and we obtain a consistent scaling in the small $\rho$ behaviour.  Despite this fact, the algorithmic bounds of different relevant works for the same problem \cite{LOFFL,FAN,JIN} seem, at first sight, to not agree with our findings. In particular \cite{LOFFL} proves rigorously the existence of an algorithm that achieves minimax rate under the BBP threshold for clustering of two-classes sparse GMM. This apparent inconsistency is related to different sparsity regimes analyzed. Indeed, here we investigate the extensive sparsity regime, i.e. $\rho = O(1)$, while the guarantees for the efficient algorithms working under the BBP threshold require a very high sparsity level, i.e. $\rho = o(1)$. This difference is crucial.
In Fig.~\ref{fig:dtr} we illustrate it by considering  the performance of two popular algorithms for this problem: \emph{Diagonal Thresholding} (DT)  \cite{JLU09} and SPCA. For extremely large sparsity, i.e. $s = O(1)$, these algorithms indeed provide estimators with positive correlation with the true classes below the BBP threshold! However, as soon as we increase the density of non-zero components the performance strongly deteriorates. In fact, the transition to random chance performance takes place as the number of non-zero components approach $s \lesssim \sqrt{n}$, in agreement with the literature \cite{JIN,LOFFL}. 
%Indeed, recalling that in our setting the norm of the two $s$-sparse cluster means is $\lambda$, it was conjectured \cite{JIN} and later proved \cite{LOFFL} using a detection-to-reduction technique that the subspace clustering problem is algorithmically hard if we have: $\lambda^2=\Omega(\sfrac{s^2}{n})$ (neglecting log-factors).

%%%%%%%%%%%%%%%%%%%%%%%%%%%%%%%%%%%%%%%%%%%%
\section{Conclusion}
\label{sec:main:concl}
In this work we considered the problem of clustering $k$ homogeneous Gaussian mixtures with sparse means. Mapping this to a low-rank matrix factorization problem, we have provided an exact asymptotic characterization of the MMSE in the high dimensional limit. The Bayes-optimal performance was compared to AMP, the best known polynomial time algorithm for this problem in the studied regime. In the large sparsity regime, we uncovered a large statistical-to-computational gap as the sparsity level grows, and unveiled the existence of a computationally \emph{hard phase}. In particular, we have shown that the SNR threshold below which recovery is statistically impossible is given by $\lambda_{\text{it}} \approx \sfrac{\sqrt{-k \rho \log{\rho}}}{\sqrt{\alpha}}$, while the one for which AMP positively correlates with the ground truth classes is given by $\lambda_{\text{alg}} \ge \sfrac{k }{\sqrt{\alpha}} $. Our result for the existence of an algorithmically hard region was compared with the existing literature for this problem, solving an apparent contradiction due to the scaling assumption of the sparsity level with the dimension of the features. We corroborated our findings with the help of  algorithms for subspace clustering such as sparse principal component analysis and diagonal thresholding. The mapping of the subspace clustering problem to a low-rank matrix factorization one is flexible and extending this work to more general scenarios is definitely interesting. 
The main limitation of the current setting is the Gaussian assumption for the noise. One natural idea to go beyond this limitation is to consider other flavours of message passing algorithms, such as vector AMP (VAMP) \cite{vamp} to study more general rotationally invariant noise distributions.
% One of the main limitations of the current setting is the Gaussian assumption for the noise, a prospective direction is to study rotationally invariant noise distribution using an extension of AMP known as Vector-AMP \cite{vamp}. 
A further prospective direction is to consider inhomegeneous mixture models, i.e. $p_c \neq \sfrac{1}{k}$ in eq.~\eqref{eq:main:sparsegmm}, in order to study the effect of unbalancedness on the statistical-to-computational gap.
%%%%%%%%%%%%%%%%%%%%%%%%%%%%%%%%%%%%%%%%%%%%
\section*{Acknowledgements}
We thank Maria Refinetti for discussions. This work started as a part of the doctoral course \emph{Statistical Physics For Optimization and Learning} taught at EPFL in spring 2021. We acknowledge funding from the ERC under the European Union’s Horizon 2020 Research and Innovation Program Grant Agreement 714608-SMiLe, and by the Swiss National Science Foundation grant SNFS OperaGOST, $200021\_200390$.
%%%%%%%%%%%%%%%%%%%%%%%%%%%%%%%%%%%%%%%%%%%%
% Bibliography
\newpage
\bibliographystyle{unsrt}
\bibliography{refs}

%%%%%%%%%%%%%%%%%%%%%%%%%%%%%%%%%%%%%%%%%%%%
%%%%%%%%%%%%%% Appendix %%%%%%%%%%%%%%%%%%%%
%%%%%%%%%%%%%%%%%%%%%%%%%%%%%%%%%%%%%%%%%%%%
\newpage
\appendix
\section*{Appendix}
%%%%%%%%%%%%%%%%%%%%%%%%%%%%%%%%%%%%%%%%%%%%
\section{Analysis of the thresholds}
\label{sec:app:potential}
\begin{figure}
\centering
     \includegraphics[width=0.75\textwidth]{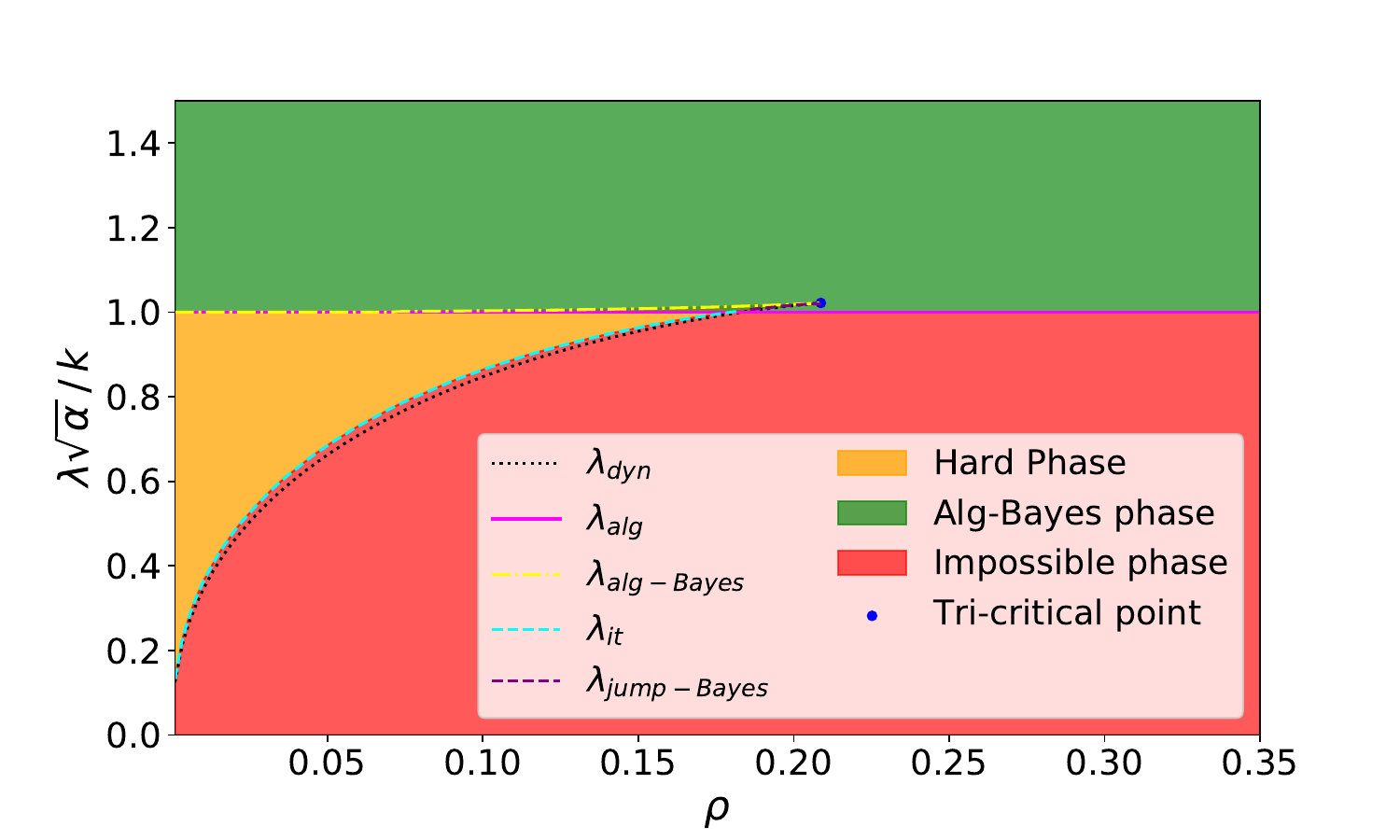}
      \caption{Enriched phase diagram for the subspace clustering of two-clusters GMM at fixed $\alpha=2$.  We plot the SNR $\lambda$ as a function of $\rho$ and we rescale the y-axis by $\sfrac{\sqrt{\alpha}}{k}$. We colour different region of the figure according to the associated phase. We introduce in the black dotted line the dynamical spinodal threshold $\lambda_{\text{dyn}}$, in the yellow dashdotted line the Bayes-algorithmical threshold $\lambda_{\text{alg-Bayes}}$ and in purple dashed line the jump-Bayes one $\lambda_{\text{jump-Bayes}}$. It is not visible, due to the choice of the axis, the \emph{easy} region, in which AMP performs better than random but not Bayes-optimally. We analyze this in  Fig.~\ref{fig:zoom_tric}.}
\label{fig:pd_full}
\end{figure}

We identified in Sec.~\ref{sec:main:thresh} different \emph{reconstruction phases} for the subspace clustering problem, characterizing completely the Bayes-optimal and algorithmical performances. We detail in this section the definition of the reconstruction phases and the consequences for the computational and statistical limits of subspace clustering. We discuss in particular an interesting link between the thresholds separating these phases and the potential function $\Phi_{\rs}$ in eq.~\eqref{eq:app:phi}.
First, we enlarge the picture on the reconstruction phases we offered in Sec.~\ref{sec:main:theory}. Along with the impossible, hard and easy phases we can define a further region. We call the \emph{Alg-Bayes} phase, the region of parameters in which the performance of AMP is, not only achieving positive correlation with the ground truth, but achieves the Bayes-optimal performance. We summarize now the complete description:
\begin{description}[wide=1pt]
    \item[Impossible phase:] There is not enough information in the data matrix $X$ handled to the statistician in order to assign cluster membership better than chance for \textit{any algorithm}, and the Bayes-optimal MMSE is not better than a random guess. Clustering (i.e. reconstruction of $\mat{U}_{\star}$ better than chance) is impossible.
    \item[Hard phase: ] The MMSE is non-trivial, and clustering is statistically possible to some extent, but the best known polynomial time algorithm, AMP, fails to correlate better than chance with the true cluster assignment $\mat{U}_{\star}$. Any polynomial-time algorithm  is conjectured to fail in this region. 
    \item[Easy phase: ] In the easy phase, not only clustering is statistically possible, but AMP is able to achieve positive correlation with $\mat{U}_{\star}$.
    \item[Alg-Bayes phase: ] In the alg-Bayes phase AMP is able to achieve Bayes-optimal positive correlation with $\mat{U}_{\star}$.
\end{description}
Following the introduction of the alg-Bayes phase, we find an enriched version of the phase diagram for the two-classes subspace clustering at fixed $\alpha$, see Fig.~\ref{fig:pd_full}. 
When we cross from one region to an other we have a \emph{phase transition}. Each phase transition is characterized by different \emph{thresholds}: values of the parameters which signal the onset of a new phase. In Fig.~\ref{fig:pd_full} we see different thresholds which were not present in the previous plot in Fig.~\ref{fig:pd}: $\{\lambda_{\text{dyn}},\lambda_{\text{alg-Bayes}},\lambda_{\text{jump-Bayes}}\}$.
In order to define these quantities, it is useful to highlight the relationship between the thresholds and the minima of the "free energy" $\Phi_{\rs}$ in eq.~\eqref{eq:app:phi}.
Let us fix  a sparsity level $1-\rho$, say $\rho = 0.05$, and move vertically on the y-axis starting from the bottom on the y-axis, i.e. $\sfrac{\lambda \sqrt{\alpha}}{k} \ll 1$. As we increase the value of the SNR we can identify the following thresholds:
\begin{itemize}[topsep=0pt,wide = 0pt]
    \item 
    $\lambda < \lambda_{\text{dyn}}$: The only minimum of eq.~\eqref{eq:res:argmin} is the trivial minimum corresponding to zero correlation $\mat{M}_{u}=0$ with $\mat{U}_{\star}$. Therefore, below this threshold reconstruction is impossible.
    \item $\lambda \in (\lambda_{\text{dyn}},\lambda_{\it})$: A second minima with higher $\Phi_{\rs}$ (i.e. a local minimum) and correlation appears, but the trivial minimum $\mat{M}_{u}=0$ is still the global one. Therefore, AMP with a uninformed initialization $\mat{M}^{t=0} = 0$ will converge to the trivial minimum, and reconstruction is only possible with a strong informed initialization. We call the threshold value for the emergence of this local minimum the \emph{dynamical spinodal} $\lambda_{\text{dyn}}$.
    
    \item $\lambda \in (\lambda_{\it},\lambda_{\text{alg}})$: 
    As the SNR is increased, the local minumum goes down in energy $\Phi_{\rs}$, and at a certain $\lambda_{\it}$, it crosses the trivial minimum. Therefore, in this interval the non-trivial minimum is the global one, while the trivial minimum becomes local. Although reconstruction is statistically possible in this region, AMP with a uninformed initialization $\mat{M}^{t=0}=0$ is stuck at the trivial minima. Therefore, in this region we enter the hard phase.
    
    \item $\lambda \in (\lambda_{\text{alg}},\lambda_{\text{alg-Bayes}})$ As the SNR is further increased, AMP with a uninformed initialization starts to achieve positive correlation with $\mat{U}_{\star}$, although strictly lower than the Bayes-optimal estimator. In terms of the potential $\Phi_{\rs}$, this corresponds to the trivial minimum continuously becoming a local maximum, and another local minimum corresponding to higher correlation continuously appearing. This new local minima coexists with the global one, which corresponds to the Bayes-optimal performance. We enter the easy phase.
    
    \item $\lambda > \lambda_{\text{alg-Bayes}}$: Finally, as the SNR is further increased the local minima disappears, and there is only one minimum with high-correlation with the signal left. In this region, AMP with a uninformed initialization achieves the same performance as the Bayes-optimal estimator. We enter the alg-Bayes phase.
\end{itemize}
\begin{figure}
\centering
     \includegraphics[width=\textwidth]{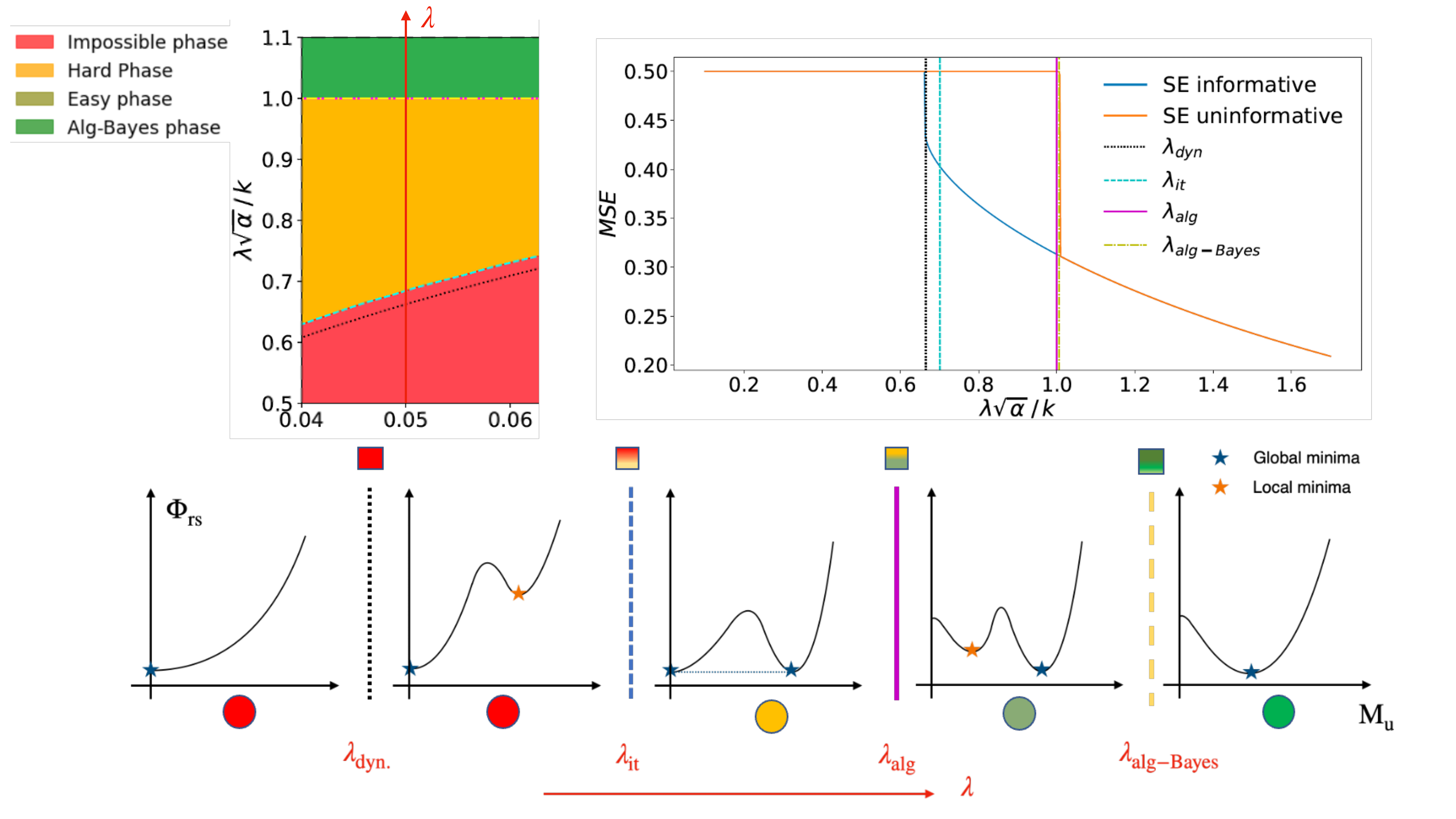}
      \caption{Evolution of the MSE and the potential $\Phi_{\rs}$  for fixed $\rho \simeq 0.05$ and $\alpha=2$. Top: We take vertical cross section of the phase diagram in Fig.~\ref{fig:pd_full} for $\rho \simeq 0.05$, as explained in the left panel. In the right panel we analyze the MSE via SE both informed and uninformed as a function of the SNR $\lambda$ and we rescale the x-axis by $\sfrac{\sqrt{\alpha}}{k}$. In vertical line we plot the different thresholds we encounter as we increase the SNR. Bottom: Cartoon plot of the minima of the potential $\Phi_{\rs}$ as we follow the cross section of the phase diagram at $\rho \simeq 0.05$. The blue star denotes the global minimum of $\Phi_{\rs}$, which corresponds to the correlation of the Bayesian-optimal estimator, while the orange dot denote local minima. We label the phase in which we are at a given stage by colouring the circle below every plot. We plot the thresholds as vertical lines separating the different subplots.}
\label{fig:pot_0.05}
\end{figure}
\begin{figure}
\centering
     \includegraphics[width=0.75\textwidth]{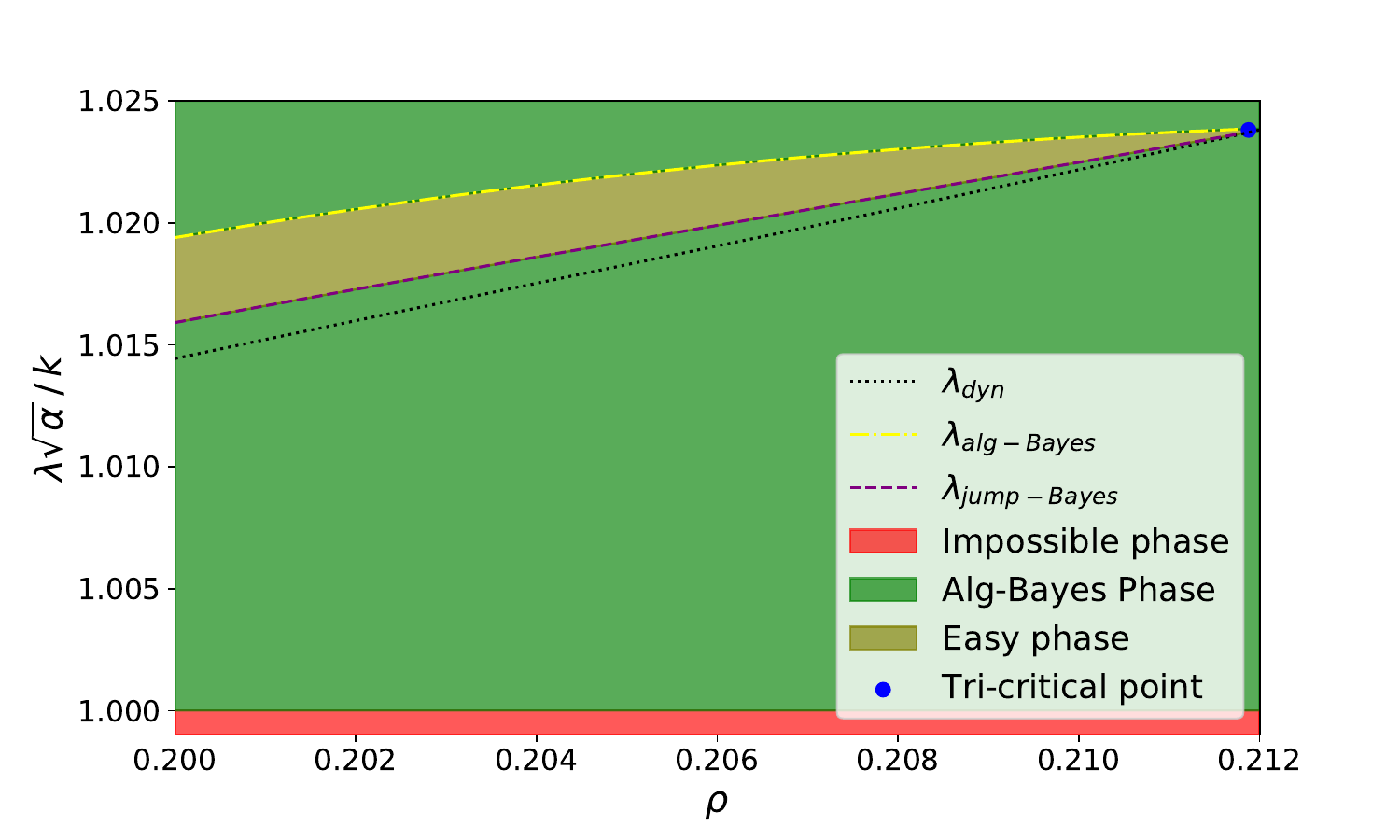}
      \caption{Zoom around the tri-critical point of Fig.~\ref{fig:pd_full}.}
\label{fig:zoom_tric}
\end{figure}

\begin{figure}
\centering
     \includegraphics[width=\textwidth]{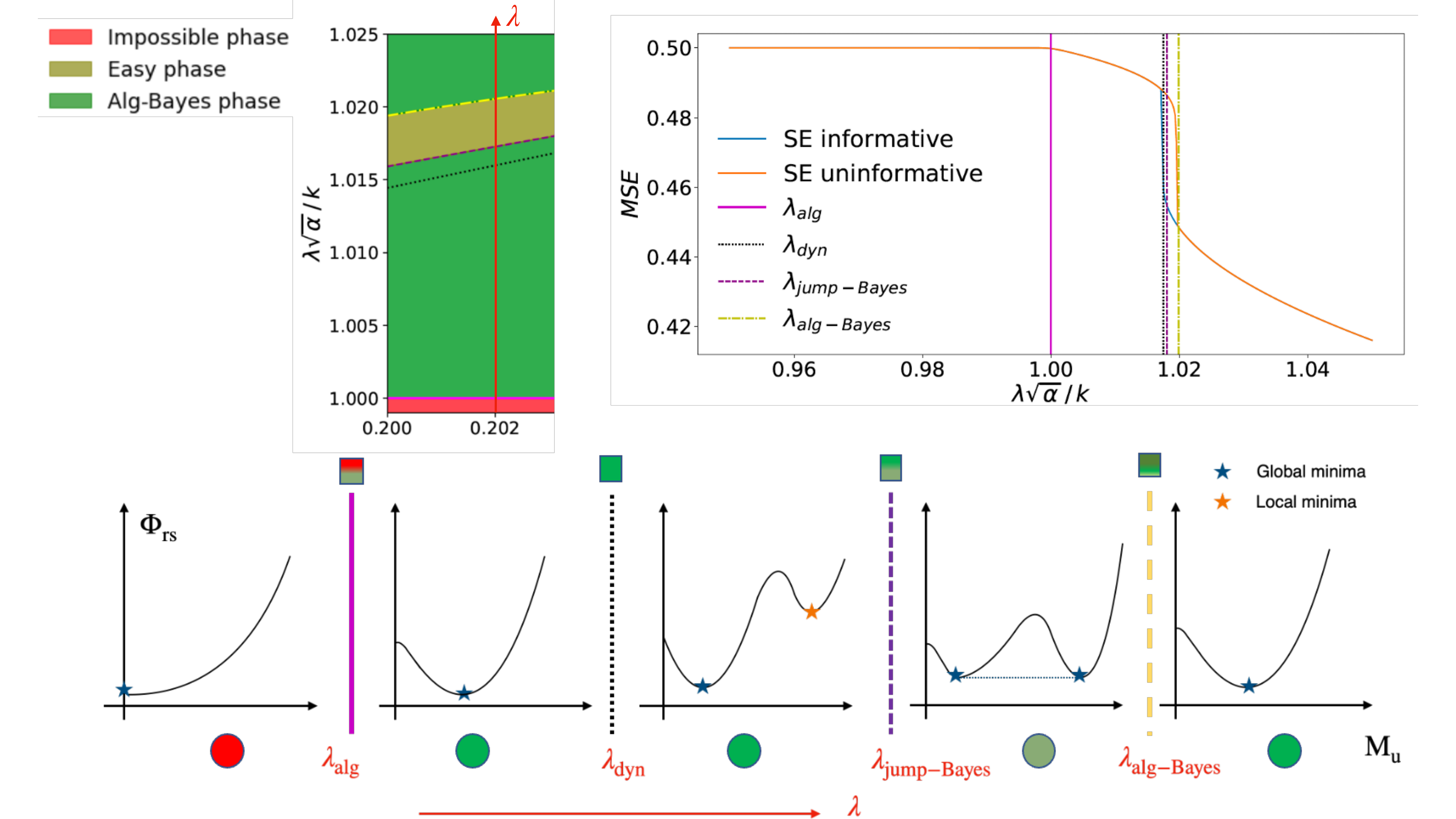}
      \caption{Evolution of the MSE and the potential $\Phi_{\rs}$  for fixed $\rho \simeq 0.202$ and $\alpha=2$. Top: We take a vertical cross section of the phase diagram in Fig.~\ref{fig:pd_full} for $\rho \simeq 0.202$, as explained in the left panel. In the right panel we analyze the MSE via SE both informed and uninformed as a function of the SNR $\lambda$ and we rescale the x-axis by $\sfrac{\sqrt{\alpha}}{k}$. In vertical line we plot the different thresholds we encounter as we increase the SNR. Bottom: Cartoon plot of the minima of the potential $\Phi_{\rs}$ as we follow the cross section of the phase diagram at $\rho \simeq 0.202$. The blue star denotes the global minimum of $\Phi_{\rs}$, which corresponds to the correlation of the Bayesian-optimal estimator, while the orange dot denote local minima. We label the phase in which we are, at a given stage, by colouring the circle below every plot. We plot the thresholds as vertical lines separating the different subplots.}
\label{fig:pot_0.2}
\end{figure}

The discussion above is summarized in Fig.~\ref{fig:pot_0.2}.
We plot toghether with the evolution of the minima of $\Phi_{\rs}$, the performance of SE with both informative and uninformative initialization to analyze the behaviour of the MSE.

The full characterization of the subspace clustering problem both from an algorithmic and statistical perspective, boils down to the analysis of the evolution of the critical points of $\Phi_{\rs}$ as we vary the meaningful parameters in the problem. 
We note from Fig.~\ref{fig:pd_full} that the performance of AMP is, when it achieves positive correlation with the ground truth, almost everywhere Bayes-optimal apart from a small region around the \emph{tri-critical} point. This point is defined - at fixed $(\alpha,k)$- as the tuple of parameters  $\left(\lambda_T(\alpha,k\right),\rho_T(\alpha,k))$, such that the "spinodal" thresholds meet, i.e. $\lambda_{\text{alg-Bayes}} = \lambda_{\text{dyn}}$. We discuss why these thresholds are called \emph{spinodals}, and how to compute them practically in Sec.~\ref{sec:app:pd}.
We can analyze the vicinity of the tri-critical point to analyze the non-trivial interplay between the easy and alg-Bayes phase, where AMP does not achieve always Bayes-optimal performance. Imagine to repeat the same steps as before considering the zoom around the tri-critical point of the phase diagram, see Fig.~\ref{fig:zoom_tric}. Fix  a sparsity level $1-\rho$, say $\rho = 0.202$, and move vertically on the y-axis starting from the bottom on the y-axis, i.e. $\sfrac{\lambda \sqrt{\alpha}}{k} \ll 1$.
As we increase the SNR we can repeat the previous analysis, obtaining now:
\begin{itemize}[topsep=0pt,wide = 0pt]
    \item 
    $\lambda < \lambda_{\text{alg}}$: The only minimum of eq.~\eqref{eq:res:argmin} is the trivial minimum corresponding to zero correlation $\mat{M}_{u}=0$ with $\mat{U}_{\star}$. Therefore, below this threshold reconstruction is impossible.
    \item $\lambda \in (\lambda_{\text{alg}},\lambda_{\text{dyn}})$: The trivial minimum becomes unstable and AMP achieves positive correlation with the ground truth. The minimum is unique and also SE with a positive initialization would end up there. The phase is alg-Bayes.
    
    \item $\lambda \in (\lambda_{\text{dyn}},\lambda_{\text{jump-bayes}})$: 
    As the SNR is increased, a new local minimum appears. The reconstruction phase is still alg-Bayes since the non-trivial minimum has higher free energy than the global one.
    
    \item $\lambda \in (\lambda_{\text{jump-Bayes}},\lambda_{\text{alg-Bayes}})$ As the SNR is further increased, the free energy of the informative minimum goes down and becomes equal to the uninformative one. We enter the easy phase, nevertheless AMP achieves positive correlation with the truth, the Bayes optimal performance is superior. The Bayes-optimal MSE have a first order phase transition at $\lambda_{\text{jump-Bayes}}$, hence the name \emph{jump-Bayes}.
    \item $\lambda > \lambda_{\text{alg-Bayes}}$: Finally, as the SNR is further increased the local minima disappears, and there is only one minimum with high-correlation with the signal left. In this region, AMP with a uninformed initialization achieves the same performance as the Bayes-optimal estimator. We enter the alg-Bayes phase.
\end{itemize}
The analysis of the evolution of the minima of $\Phi_{\rs}$ and the consequences on the MSE is done in Fig.~\ref{fig:pot_0.2}. 

%%%%%%%%%%%%%%%%%%%%%%%
\section{Building the phase diagram}
\label{sec:app:pd}
We build in this section the phase diagram for the two-classes GMM shown in Fig.~\ref{fig:pd} , explaining the steps which are easily generalizable to the general mixture case. First we note that we can simplify the model for two-mixtures GMM in eq.~\eqref{eq:model} even further by mapping it to an easier rank $k=1$ version of the matrix factorization problem. It suffices to replace the  matrices $(U,V)$ in eq.~\eqref{eq:model} by the following quantities:
\begin{align}
    \vec{u} \sim \text{Rad}(n) \in \{-1,+1\}^n  && \vec{v} \sim_{i.i.d.} \rho  \mathcal{N}(0,\mat{I}_{d})+(1-\rho)\delta_{0} \in \mathbb{R}^d
    \label{eq:app:simplified_k2}
\end{align}
The two formulations of the problem are indeed formally equivalent up to a rescaling of the parameters such that at fixed $\alpha$ the quantity $\sfrac{\lambda}{k}$ is the same in two settings. The mapping simplify significantly the computation. First, in order to compute the Bayes-optimal performance in eq.~\eqref{eq:res:mmse}, we must compute the \emph{partition functions} $Z_{u/v}$ in eqs.~\eqref{eq:main:subspace_Zs1},\eqref{eq:main:subspace_Zs2} for the new simplified model. We shall exploit the following general relation, as a function of the prior distribution on  $(\mat{U},\mat{V})$: 
\begin{align}
    Z_{u/v}(\mat{A},\vec{b}) = \EX_{\vec{x}\sim P_{u/v}}\left[\exp \left(-\vec{b}^{\top}\vec{x} + \frac{\vec{x}^{\top}\mat{A}\vec{x}}{2}\right)\right]
\end{align}
\noindent thus exploiting the explicit expression of the prior in eq.~\eqref{eq:app:simplified_k2} we obtain:
\begin{align}
    Z_u (A,b)=  e^{-\frac{A}{2}}\cosh{b} && 
    Z_v (A,b)= 1-\rho + \frac{\rho}{\sqrt{1+A}}\exp \left(\frac{b^2}{2(1+A)}\right)
\end{align}
We study the computational limits for the subspace clustering problem deriving the
associated AMP to this simplified low-rank matrix factorization problem.  We have to compute the \emph{denoising functions} for the simplified model, written in eqs.~\eqref{eq:main:denoiser_u},\eqref{eq:main:denoiser_v} for the general rank case. We shall exploit the general formula relating them with the partition functions computed above:
\begin{align}
    \vec{\eta}_{u/v}(\mat{A},\vec{b}) = \partial_{\vec{b}} \log Z_{u/v}(\mat{A},\vec{b})
\end{align}
\noindent thus using the prior for the simplified $k=1$ model in eq.~\eqref{eq:app:simplified_k2} we obtain:
\begin{align}
    \eta_{u}(A,b) = \tanh(b) &&
    \eta_v(A,b) = \frac{\rho b}{1+A}\frac{1}{\rho +(1-\rho)\sqrt{1+A}e^{-\frac{b^2}{2(1+A)}}}
\end{align}
\noindent where now $(A,b) \in \mathbb{R}^2$. We can write at this point the SE equations for the overlaps $(m_u^t,m_v^t)$ which are now scalar variables. Let us introduce the following notation:
\begin{align}
     m_u^{t+1} &= \EX_{u_*\sim P_u,\xi \sim \mathcal{N}(0,1)}\left[\eta_u\left(\frac{\lambda m_v^t}{\rho},  \frac{\lambda m_v^t u_*}{\rho} + \sqrt{\frac{\lambda m_v^t}{\rho}} \xi \right)u_*\right] \coloneqq \mathcal{U}(\sfrac{\lambda m_v^t}{\rho})
      \label{eq:app:se_eq1}\\
     m_v^{t+1} &= \EX_{v_*\sim P_v,\xi \sim \mathcal{N}(0,1)}\left[\eta_v\left(\frac{\alpha \lambda m_u^t}{\rho}, \frac{\alpha \lambda m_u^t v_*}{\rho} + \sqrt{\frac{\alpha \lambda m_u^t}{\rho}} \xi\right)v_*\right] \coloneqq \mathcal{V}(\sfrac{\alpha \lambda m_u^t}{\rho})
      \label{eq:app:se_eq2}
 \end{align}
The perturbative method we presented in Sec~\ref{sec:main:k2}  would pinpoint easily the expected algorithmical threshold $\lambda_{\text{alg}} = \sfrac{1}{\sqrt{\alpha}}$. Despite this, as we discussed in Sec.~\ref{sec:main:k2}, it would not guarantee the presence of an algorithmically hard region since the number of cluster $k<k_{\text{hard}}$. 
\begin{figure}[H]
\centering
     \includegraphics[width= \textwidth]{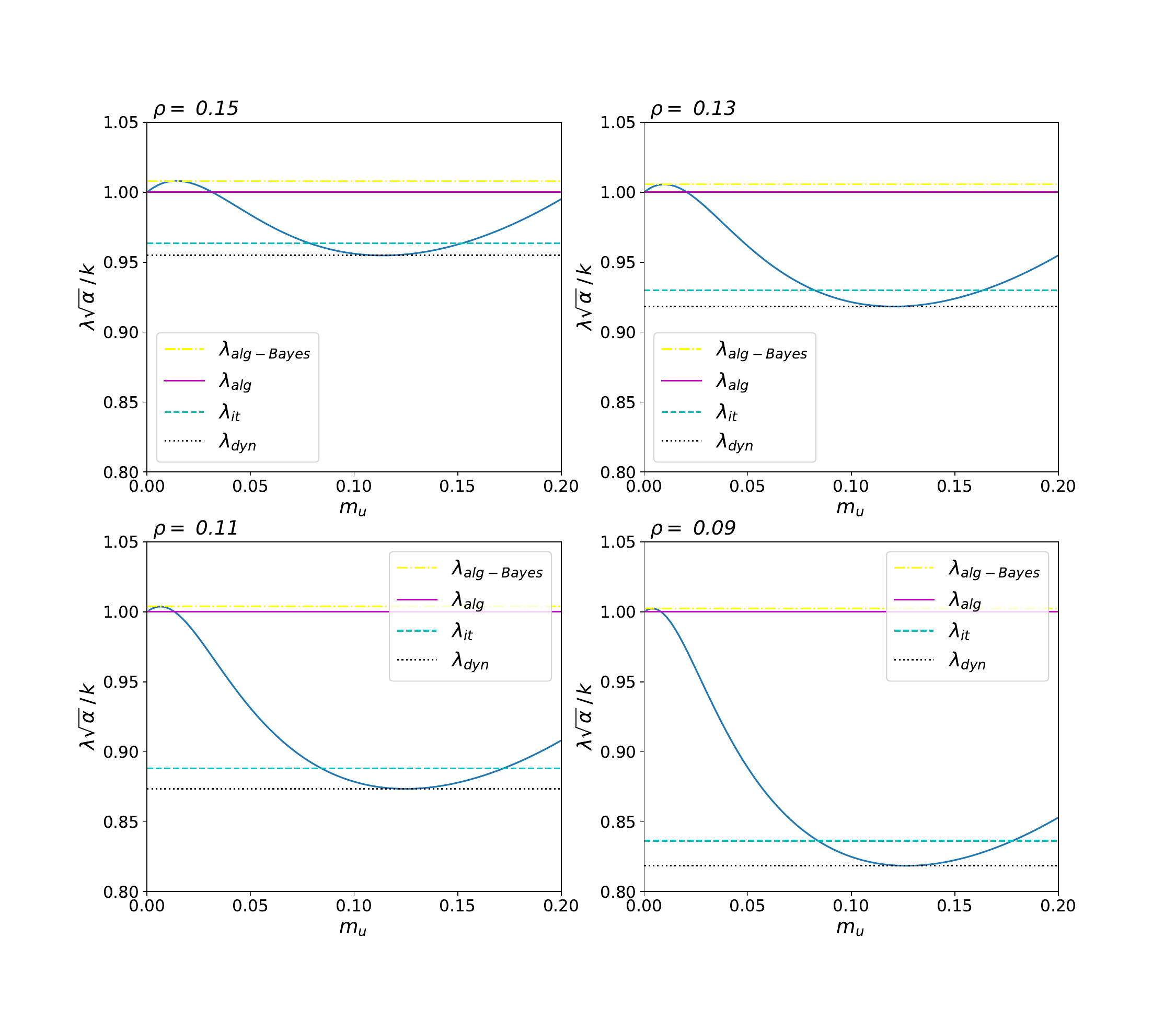}
      \caption{Evolution of the SNR $\lambda$ as a function of $m_u$ for different $\rho \in \{ 0.09,0.11,0.13,0.15 \}$. We rescale the y-axis by $\sfrac{\sqrt{\alpha}}{k}$. We plot the different transitions $\{\lambda_{\text{dyn}},\lambda_{\text{it}},\lambda_{\text{alg},},\lambda_{\text{alg-Bayes}}\}$ for each different sparsity level. It is visually clear that the hard phase becomes bigger and bigger as the sparsity grows, while the gap between dynamical and IT threshold closes.} 
\label{fig:spin}
\end{figure}
We have to resort to numerical computations for finding the exact values of the thresholds since also in this simple case we do not have a closed-form update for the iterates $(m_u^t,m_v^t)$, although we will treat them analytically in App.~\ref{sec:app:large_spars}.
We keep in mind the picture in Figs.~(\ref{fig:pot_0.05},\ref{fig:pot_0.2}) and compute the different thresholds there defined. Let us consider first $\lambda_{\text{it}}$, the IT threshold, defined as the SNR at which the problem becomes statistically possible. We see in Fig.~\ref{fig:pot_0.05} that it coincides with the SNR level at which the free energy of the two minima (if present) are equal. We analyze for simplicity sparsity levels in which $\lambda_{\text{dyn}} < \lambda_{\text{alg}}$, i.e. we refer to Fig.~\ref{fig:pot_0.05}, otherwise the criterion above would define equivalently $\lambda_{\text{bayes-Jump}}$. We compute the difference of free energy $\Delta \Phi$ between the two minima introducing the path $\gamma: \mathbb{R} \to \mathbb{R}^2$ which follows the state evolution equations: $$
\gamma(t) = (t, \mathcal{V}(\alpha \lambda t))$$
We can use the fundamental theorem of calculus to obtain the difference of free energy between the trivial fixed point $(m_u,m_v)=(0,0)$ and a non-trivial one $(m_u,m_v)=\left(x,\mathcal{V}\big(\alpha \lambda(x) x)\right)$ at overlap $m_u = x $ as follows:
\begin{align}
    \Delta \Phi (x) 
    = \int_0^x \dd q \frac{d \Phi}{dt}\big (m_u(t) = q, m_v(t) = \mathcal{V}(\alpha \lambda(x) q);\lambda(x),\rho,\alpha \big )
    \label{eq:it_eq}
\end{align}
\noindent where we introduced $\lambda(x)$ as the SNR needed in order to have at fixed $(\rho,\alpha)$ an overlap $m_u=x$ defined by the self-consistent equation: 
\begin{align}
    x = \mathcal{U}\big(\lambda(x)\mathcal{V}(\alpha \lambda(x) x)\big) 
    \label{eq:self_cons_eq_snr}
\end{align}
Plugging in the expression of the derivative we identify the IT threshold $\lambda_{\text{it}}$ as the minimal SNR such that the following equation is satisfied:
\begin{align}
     \int_0^{x(\lambda_{\textit{it}})} dq \mathcal{V}'(\alpha \lambda_{\textit{it}} q) [q - \mathcal{U}(\lambda_{\textit{it}} \mathcal{V}(\alpha \lambda_{\textit{it}} q))] = 0
\end{align}
Let us consider now the Bayes-algorithmical and dynamical thresholds, always referring to their pictorial representation in Figs.~(\ref{fig:pot_0.05},\ref{fig:pot_0.2}). From a practical standpoint they are stationary point of the the function $\lambda(m_u)$, solution of eq.~\eqref{eq:self_cons_eq_snr}. A reader with some statistical physics background may recognize a parallel with the theory of real gases. The curve $\lambda(m_u)$, exactly as the Pressure-Volume curve $p(v)$ for real gases, is composed of two branches called \emph{stable} and \emph{unstable} branch defined from the value of the derivative $\sfrac{\partial \lambda}{\partial m}$ (resp. $\sfrac{\partial p}{\partial v}$). The operative definition of these thresholds as critical points of the curve $\lambda({m_u})$ allows us to easily compute them numerically, see Fig.~\ref{fig:spin} to observe their evolution as a function of the sparsity level. We see that as we increase the sparsity, i.e. decrease $\rho$, the statistical-to-computation gap, measured visually by the distance between the two critical points, increase. Moreover the IT threshold collapse with the dynamical one. The same happens with the Bayes-algorithmic threshold, approaching $\lambda_{\text{alg}} \approx 1$.

%%%%%%%%%%%%%%%%%%%%%%%
%%%%%%%%%%%%%%%%%%%%%%%%%
\section{Scaling behaviour at large sparsity}
\label{sec:app:large_spars}
\begin{figure}
\centering
     \includegraphics[width= \textwidth]{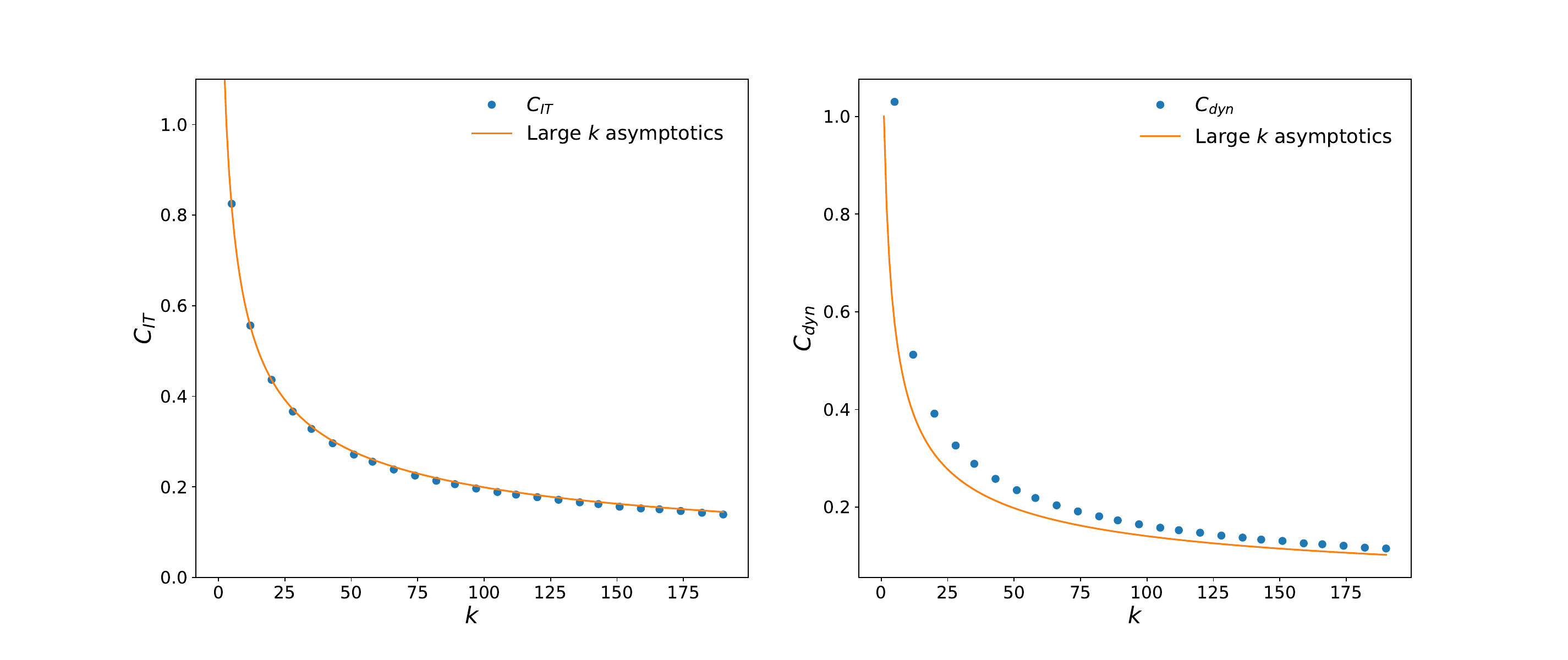} 
      \caption{\emph{Left}: Comparison of the threshold coefficients $(C_{\text{it}}, C_{\text{dyn}})$ (dots) with their high rank asymptotic expression (solid line).   }
\label{fig:C_trans}
\end{figure}

We do not have in general an analytical expression for the iterates of the SE equation $(m_u^t,m_v^t)$ appearing in eq.~\eqref{eq:se_simple}, thus we introduced in Sec.~\ref{sec:main:small} a change of variables that allows us to approach analytically the problem in the large sparsity regime. We study in this section the consequences of this scaling assumption. Let us rewrite it here:
\begin{align}
    m_u = \tilde{m}_u \sqrt{\frac{-\rho  \log{\rho}}{\alpha}} &&
    m_v = \tilde{m}_v \rho  && 
    \lambda = C(k) k \sqrt{\frac{-\rho \log{\rho}}{\alpha}}
    \label{eq:ansatz_app}
\end{align}
Consider the update of the parameter $m_v$ under this parametrization:
\begin{align}
    m_v = \rho \tilde{m}_v  = f_v^{(k)}\big(-\tilde{m}_u C(k) \log{\rho}\big) 
\end{align}
We can rewrite the right hand side in the following way:
\begin{align}
    \rho \tilde{m}_v = \rho \frac{-\tilde{m}_u C(k) \log{\rho}}{(k-\tilde{m}_u C(k) \log{\rho})}  \int_{0}^{+ \infty}\frac{S_{k-1}}{(2 \pi)^{\frac{k}{2}}}  \frac{\rho \xi^{k+1}e^{-\sfrac{\xi^2}{2}}}{\rho + (1-\rho)(\frac{k-\tilde{m}_u C(k) \log{\rho}}{k})^{\frac{k}{2}}\rho^{\sfrac{\tilde{m}_u C(k)\xi^2}{2}}} \, \dd\xi
\end{align}
\noindent where $S_{k-1}(1)$ is the surface of the $k-$dimensional unitary hypersphere.
Working in the small $\rho$ limit allows us to exploit a concentration in measure over which we integrate on the right hand side. The exponent of $\rho$ in the denominator, i.e. $\frac{\tilde{m}_u C(k) \xi^2}{2}$, will determine the large sparsity behaviour. If the exponent is greater than one, $\rho^{\sfrac{\tilde{m}_u C(k) \xi^2}{2}}$ will go to zero, otherwise it will diverge in the limit $\rho \to 0$. Thus we obtain:
\begin{align}
    m_v \approx \rho \int_{0}^{+\infty}\frac{S_{k-1}(1)}{k (2 \pi)^{\sfrac{k}{2}}}\xi^{k+1}e^{-\sfrac{\xi^2}{2}} \Theta \left(\frac{\tilde{m}_u C(k)}{2}\xi^2 - 1 \right) \coloneqq \rho T_k\left(\tilde{m}_u C(k)\right)
    \label{eq:small_rho_exp}
\end{align}
\noindent where we introduced $\Theta(x)$ as the Heavyside theta. Plugging in the above expression into the equation defining $m_u^t$ we have:
\begin{align}
    m_u = f_u^{(k)}\left(\lambda \rho T_k\left(\tilde{m}_u C(k)\right)\right)
\end{align}
\noindent thus approximating for small $\rho$ the function $f_u^{(k)}$, expressing everything in terms of $m_u$ and plugging in the scaling ansatz for $m_u$ in eq.~\eqref{eq:ansatz_app}, we obtain the simplified SE in the large sparsity regime:
\begin{align}
    \Tilde{m}_u = C(k) T_k\left(\tilde{m}_u C(k)\right) \label{eq:small_rho_se}
\end{align}
We can repeat the analysis done in the previous appendix to find the thresholds in this limit. The condition defining the IT threshold $\lambda_{\it}$ written in eq.~\eqref{eq:it_eq} simplifies to:
\begin{align}
    \int_0^{\Tilde{x}} d\tilde{q}\, T_k^{\prime}\left(C_k(\tilde{x})\tilde{q}\right)\tilde{q}= \int_0^{\Tilde{x}} d\tilde{q} \, T_k^{\prime}\left(C_k(\tilde{x})\tilde{q}\right)C_k(\tilde{x})T_k\left(C_k(\tilde{x})\tilde{q}\right)
    \label{eq:app:C_it}
\end{align}
\noindent where we defined $C_k(\tilde{x})$ as the value of the coefficient $C(k)$, solution of eq.~\eqref{eq:small_rho_se} when the overlap is fixed at $m_u = \sqrt{\frac{-\rho  \log{\rho}}{\alpha}} \tilde{x}$. This task is much easier to solve. Likewise, the computation of the dynamical spinodal simplifies greatly. We need to find the minimum SNR such that eq.~\eqref{eq:small_rho_se} has a non trivial solution. By introducing the auxiliary variable $y = C_k(y) \tilde{m}$ we rewrite eq.~\eqref{eq:small_rho_se} as:
\begin{align}
    C_k^2(y) = \frac{y}{T_k(y)}
\end{align}
\noindent thus the minimal SNR to obtain a non trivial solution, defined by the coefficient $C_{\text{dyn}}(k)$, to obtain a non trivial solution of the equation above is given by:
\begin{align}
    C_{\text{dyn}}(k) = \min_{y\in \mathbb{R}^+} \sqrt{\frac{y}{T_k(y)}} 
    \label{eq:app:Cdyn}
\end{align}
At this stage we still need to resort to numerical inspection in order to find the coefficient $C_{\it},C_{\text{dyn}}$, but we can investigate analytically the large $k$ behaviour. By considering the leading order of the function $T_k(\cdot)$, one can see that $T_k(z) \approx \Theta\left(z-\frac{2}{k+1}\right)$. By plugging in this expression into the definition of the coefficients $\left(C_{\it}(k),C_{\text{dyn}}(k)\right)$ in eqs.~\eqref{eq:app:C_it},\eqref{eq:app:Cdyn} we obtain the following asymptotic result:
\begin{align}
    C_{\it}(k) \approx \sqrt{\frac{4}{k+1}} \qquad 
    C_{\text{dyn}}(k) \approx \sqrt{\frac{2}{k+1}}
    \label{eq:small_rho_C_trans}
\end{align}
\noindent thus plugging them into the scaling assumption in eq.~\eqref{eq:ansatz_app} we obtain the following scaling for the thresholds:
\begin{align}
    \lambda_{\text{it}} \approx \sqrt{ \frac{4k^2}{k+1}} \sqrt{\frac{-\rho \log{\rho}}{\alpha}}\qquad \lambda_{\text{dyn}} \approx \sqrt{ \frac{2k^2}{k+1} }\sqrt{\frac{-\rho \log{\rho}}{\alpha}} 
\end{align}
The evolution of the coefficients $C_{\text{it}},C_{\text{dyn}}$ as a function of the number of clusters is summarized in Fig.~\ref{fig:C_trans}.
We see in Fig.~\ref{fig:C_trans} that the large rank expansion is quite accurate also at moderate $k$, especially for the IT threshold. The easy phase and the alg-Bayes phase merge, hence we will not analyze the distinction between $\lambda_{\text{alg}}$ and $\lambda_{\text{alg-Bayes}}$ in this regime.
%%%%%%%%%%%%%%%%%%%%%
\section{Details on numerical simulations}
\label{sec:app:numerics}
We discuss in this section the details behind the numerical simulations presented in Sec.~\ref{sec:main:thresh}. The code is available in the  \href{https://github.com/lucpoisson/SubspaceClustering}{Github repository} associated to the manuscript. First we stress an important point on the convergence of low-rAMP algorithms. Increasing the sparsity of the problem, i.e. decreasing $\rho$, the convergence of AMP becomes more difficult. In order to solve this problem is useful to implement \emph{damping} to stabilize the iteration, see the modified AMP in Algorithm~\ref{alg:lowramp_damp}. 
\begin{algorithm}[bt]
   \otherlabel{alg:lowramp_damp}{2}
   \caption{low-rAMP with damping}
\begin{algorithmic}
   \STATE {\bfseries Input:} Data $\mat{X}\in\mathbb{R}^{d\times n}$
   \STATE Initialize $\hat{\vec{v}}_{i}^{t=0}, \hat{\vec{u}}_{\nu}^{t=0}\sim\mathcal{N}(\vec{0}_{k},\epsilon\mat{I}_{k})$, $\hat{\sigma}_{u, \nu}^{t=0} = \mat{0}_{k\times k}$, $\hat{\sigma}_{v,i}^{t=0} = \mat{0}_{k\times k}$.
  
   \FOR{$t\leq t_{\text{max}}$}
   \STATE $\mat{A}_{u}^{\text{tmp}} = \frac{\lambda}{s} \left(\hat{\mat{U}}^t\right)^{\top} \hat{\mat{U}}, \qquad A_{v}^{\text{tmp}} = \frac{\lambda}{s}\left(\hat{\mat{V}}^t\right)^{\top} \hat{\mat{V}}$ 
   \STATE $\mat{B}_{v}^{\text{tmp}} = \sqrt{\frac{\lambda}{s}}X\hat{\mat{U}}^t-\frac{\lambda}{s}\sum\limits_{\nu=1}^{n}\sigma^{t}_{u, \nu}\hat{\mat{V}}^{t-1}$, \quad  $\mat{B}_{u}^{\text{tmp}} = \sqrt{\frac{\lambda}{s}}X^{\top}V-\frac{\lambda}{s}\sum\limits_{i=1}^{d}\sigma^{t}_{v, i}\hat{\mat{U}}^{t-1}$
   \STATE Damping step with damping coefficient $\gamma$:\\ 
   $\mat{A}_u^t = (1-\gamma) A_u^{\text{tmp}} + \gamma A_u^{t-1}  \qquad \mat{A}_v^t = (1-\gamma) A_v^{\text{tmp}} + \gamma A_v^{t-1} $  \\
   $\mat{B}_u^t = (1-\gamma) B_u^{\text{tmp}} + \gamma B_u^{t-1}  \qquad \mat{B}_v^t = (1-\gamma) B_v^{\text{tmp}} + \gamma B_v^{t-1} $ \\
   \STATE Take $\{\vec{b}^t_{v,i} \in \mathbb{R}^k\}_{i=1}^d, \{\vec{b}^t_{u,\nu}\in \mathbb{R}^k\}_{\nu = 1}^n$ rows of $\mat{B}^t_{v},\mat{B}^t_u$
    \STATE $\hat{\vec{v}}_{i}^{t+1} = \eta_{v}(\mat{A}_{v}^{t}, \vec{b}_{v,i}^{t})$, \qquad $\hat{\vec{u}}_{\nu}^{t+1} = \eta_{u}(\mat{A}_{u}^{t}, \vec{b}_{u,\nu}^{t})$
   \STATE $\hat{\sigma}_{v,i}^{t+1} = \partial_{\vec{b}}\eta_{v}(\mat{A}_{v}^{t}, \vec{b}_{v,i}^{t})$, \qquad  $\hat{\sigma}_{u,\nu}^{t+1} = \partial_{\vec{b}}\eta_{u}(\mat{A}_{u}^{t}, \vec{b}_{v,\nu}^{t})$
   \STATE Here $\hat{\mat{U}}^t \in \mathbb{R}^{n\times k},\hat{\mat{V}}^t \in \mathbb{R}^{d\times k},\mat{B}_u^t \in \mathbb{R}^{n\times k}, \mat{B}_v^t \in \mathbb{R}^{d\times k},\mat{A}_u^t \in \mathbb{R}^{k\times k}, \mat{A}_v^t \in \mathbb{R}^{k\times k}$
   \ENDFOR
   \STATE {\bfseries Return:} Estimators $\hat{\vec{v}}_{\amp,i}, \hat{\vec{u}}_{\amp,\nu}\in\mathbb{R}^{k}, \hat{\sigma}_{u, \nu}, \hat{\sigma}_{v, i}\in\mathbb{R}^{k\times k}$
\end{algorithmic}
\end{algorithm}
We compared the performance of AMP with different popular algorithm in the literature. The first general-purpose algorithm we considered for subspace clustering is a modification of the sparse PCA algorithm (SPCA). Let us consider a data matrix $Y \in \mathbb{R}^{n \times d}$, where as in our notation $n$ is the number of samples and $d$ is the feature dimension. In the SPCA problem, the statistician wants to find directions in the space which maximize the variance of our dataset by constraining the cardinality of the new basis vectors. In vanilla PCA instead we try to find directions, called principal components $\{\hat{\vec{e}}_m\}_{m=1}^d$, which maximize the variance not caring if they will be given by linear combination of all the features of our problem: $\hat{\vec{e}}_m = \sum_{i=1}^d \alpha_i^{(m)} \vec{e}_i$, where we called $\{\vec{e}_i\}_{i=1}^d$ the canonical basis vectors. In SPCA we want that some of the coefficients $\alpha_i^{(m)}$ (called "loadings" in the literature) to be zero, favouring interpretability of the optimal estimator. By formulating in a variational way the problem the sparsity of the estimator is enhanced using LASSO regularization. We write the pseudocode for the program we used in the two-class subspace clustering problem in Algorithm.~\ref{alg:spca}. The unregularized problem, i.e. $\Gamma=0$, is equivalent to vanilla PCA. The comparison of the performances of the two spectral algorithms has been done in Fig.~\ref{fig:mses} and we have a clear advantage in imposing the cardinality constraint as the sparsity level increase. We considered in the sub-extensive sparsity regime in Sec.~\ref{sec:main:small} the Diagonal Thresholding algorithm (DT). The main idea is to search for spatial directions with the largest variance, and threshold the sample covariance matrix accordingly, hence the name Diagonal Thresholding.
%The algorithm is not suited to work at moderate sparsity, indeed if we add DT performance to Fig.~\ref{fig:mses} one has the situation depicted in Fig.~\ref{fig:mses+}. We clearly see it performs poorly with respect with the spectral algorithm considered before. 
    The pseudocode for the algorithm we used in the two-classes subspace clustering is given in Algorithm.~\ref{alg:dt}.
\begin{algorithm}[bt]
   \otherlabel{alg:spca}{3}
   \caption{SPCA}
\begin{algorithmic}
   \STATE {\bfseries Input:} Data $\mat{Y}\in\mathbb{R}^{n\times d}$
   \STATE Initialize $\Delta_{\text{sparsity}}=1, \Gamma = 10^{-3}$
   \WHILE{$|\Delta_{\text{sparsity}}| \ge 1$}
   {
   \STATE Solve variational problem: 
   $ (\hat{C},\hat{D}) = \underset{\mat{C}\in\mathbb{R}^{n}, \mat{D}\in \mathbb{R}^d}{\argmin}~\{ \norm{\mat{Y} - \mat{C}\mat{D}^T}_F + \Gamma \norm{\mat{D}}_1 \} $
   \STATE Compute first sparse principal component $\mat{D}$
   \STATE Compute the estimated sparsity $\hat{s} = \sum_{i=1}^d (1-\delta_{\hat{v}_i,0})$
   \STATE Compute sparsity mismatch $\Delta_{\text{sparsity}} = \rho d - \hat{s}$
   \STATE If $\Delta_{\text{sparsity}} < 0 $ decrease $\Gamma$, otherwise increase it
   }
   \ENDWHILE
   \STATE Project the data matrix onto the first sparse principal component: $\mat{P} = \mat{Y} \mat{D} \in \mathbb{R}^n$
   \STATE Compute cluster membership assignment: $\hat{\mat{U}} = \text{sign}(\mat{P})$
   \STATE {\bfseries Return:} $\text{MSE}(\hat{\mat{U}})$
\end{algorithmic}
\end{algorithm}

\begin{algorithm}[bt]
   \otherlabel{alg:dt}{4}
   \caption{Diagonal Thresholding}
\begin{algorithmic}
   \STATE {\bfseries Input:} Data $\mat{Y}\in\mathbb{R}^{n\times d}$
  \STATE Compute the sample covariance matrix $\hat{\mat{K}} = \frac{1}{n}\sum_{\nu = 1}^n \vec{y}_{\nu}\vec{y}_{\nu}^{\top}$
  \STATE Find the  directions with the $s$ largest variance, with $s = \floor{\rho d}$. 
  \STATE Call the subset of indices corresponding to the directions above $\mathcal{S}$. 
  \STATE Create $\Tilde{\mat{K}}$:
  \begin{align*}
    \Tilde{\mat{K}}_{ij} =
      \begin{cases}
          \hat{\mat{K}}_{ij}  &\text{if} \quad (i,j) \in \mathcal{S}   \\
          0 &\text{otherwise}
      \end{cases}
  \end{align*}
   \STATE Compute the largest eigenvector of the thresholded matrix $\Tilde{\mat{K}}$ and call it $\hat{\vec{v}}$.
   \STATE Project the data matrix onto the first sparse principal component: $\vec{p} = \mat{Y} \hat{\vec{v}} \in \mathbb{R}^n$
   \STATE Compute cluster membership assignment: $\hat{\vec{u}} = \text{sign}(\vec{p})$
   \STATE {\bfseries Return:} $\text{MSE}(\hat{\vec{u}})$
\end{algorithmic}
\end{algorithm}

% \begin{figure}
% \centering
%      \includegraphics[width= \textwidth]{./figs/new_mse_mod.pdf} 
%       \caption{
%       Modification of Fig.~\ref{fig:mses} adding the MSE of DT vs the 
%       %We compare the MSE of SE informed and uninformed with AMP, SPCA, PCA and DT. We plot the MSE as a function of the SNR $\lambda$ and we rescale the x-axis by $\sqrt{\alpha}$. For each algorithm considered the error bars are built using the standard deviation over fifty runs with parameters $(n=8000, d=4000)$, i.e. $\alpha=2$. We plot in vertical line the theoretical values for the Information-Theoretic threshold $\lambda_{\text{it}}$ (dashed cyan line), and the  algorithmic threshold $\lambda_{\text{alg}}$ (dotted line in magenta). The theoretical values coincide with the experimental one. The SE with uninformed initialization follows AMP as expected. \emph{Left}: The sparsity is fixed with parameter $\rho= 18 \%$. Both SPCA, PCA and DT have a worse performance with respect to AMP and in this sparsity regime we have only a marginal advantage by using SPCA with respect to PCA.  \emph{Right}: The sparsity is fixed with parameter $\rho= 5 \%$. Increasing the sparsity level the width of the algorithmically hard phase becomes bigger and the performance of SPCA becomes clearly superior to the PCA one.
%       }
% \label{fig:mses+}
% \end{figure}
%%%%%%%%%%%%%%%%%%%%%%%%%%%%%%%%%%%%%%%%%%%%%
\end{document}